\documentclass[journal]{IEEEtran}
\ifCLASSINFOpdf
\else
\fi
\usepackage{amsmath} 
\usepackage{amsfonts,amssymb}
\usepackage{url}
\usepackage{bm}
\usepackage{mathrsfs}
\usepackage[lined,boxed,commentsnumbered,ruled,linesnumbered]{algorithm2e} 
\usepackage{amsmath}
\usepackage{amssymb}
\usepackage{graphicx}
\usepackage{subfigure}
\usepackage{color}
\usepackage{cite}
\usepackage{booktabs,caption}
\usepackage[flushleft]{threeparttable}
\begin{document}

\title{Learning Decentralized Traffic Signal Controllers with Multi-Agent Graph Reinforcement Learning}
\author{
\IEEEauthorblockN{Yao~Zhang, Zhiwen~Yu,~\IEEEmembership{Senior Member,~IEEE}, Jun~Zhang,~\IEEEmembership{Fellow,~IEEE}, Liang~Wang, Tom~H.~Luan,~\IEEEmembership{Senior Member,~IEEE}, Bin~Guo,~\IEEEmembership{Senior Member,~IEEE}, Chau~Yuen,~\IEEEmembership{Fellow,~IEEE}}
\thanks{Y. Zhang, L. Wang, and B. Guo are with the School of Computer Science, Northwestern Polytechnical University, Xi'an, China.}
\thanks{{Z. Yu is with Harbin Engineering University, and also with the School of Computer Science, Northwestern Polytechnical University, Xi'an, China.}}
\thanks{J. Zhang is with the Department of Electronic and Computer Engineering, Hong Kong University of Science and Technology, Hong Kong. }
\thanks{T. H. Luan is with the School of Cyber Engineering, Xidian University,
Xi'an, China. }
\thanks{C. Yuen is with the School of Electrical and Electronics Engineering, Nanyang Technological University, Singapore.}
\thanks{J. Zhang is the corresponding author, Email: eejzhang@ust.hk.
}
}


%
 
 
\maketitle

\begin{abstract}

 
This {paper considers optimal traffic signal control in smart cities, which has been taken as a complex networked system control problem.} Given the interacting dynamics among traffic lights and road networks, attaining controller adaptivity and scalability stands out as a primary challenge.
{Capturing the spatial-temporal correlation among traffic lights under the framework of Multi-Agent Reinforcement Learning (MARL) is a promising solution. Nevertheless, existing MARL algorithms ignore effective information aggregation which is fundamental for improving the learning capacity of decentralized agents.}
In this paper, we design a new decentralized control architecture with improved environmental observability to capture {the spatial-temporal correlation. 
Specifically, we first develop a \textit{topology-aware information aggregation} strategy to extract correlation-related information from unstructured data gathered in the road network}. Particularly, we transfer the road network topology into a graph shift operator by forming a diffusion process on the topology, which subsequently facilitates the construction of graph signals. {A diffusion convolution module is developed, forming a new MARL algorithm, which endows agents with the capabilities of graph learning}. Extensive experiments based on both synthetic and real-world datasets verify that our proposal outperforms existing decentralized algorithms.
  





 
\end{abstract}


\IEEEpeerreviewmaketitle

\section{Introduction}

With the growth of population and urbanization, car parc is steadily increasing all over the world. By $2021$, the car parc of China has been raised over $1.9$ times as against ten years ago \cite{carparcChina}. As a consequence, {traffic congestion becomes a severe social problem in metropolises, inevitably incurring severe Carbon Dioxide (CO2) emissions and traffic crashes. Controlling traffic signals adaptively is a straightforward way to reduce traffic congestion.
However, it is oftentimes difficult for control policies to be adaptive and scaled up in complex traffic environments because of the dynamic correlation among traffic lights. 
To address such difficulties, traditional methods mainly focus on heuristic decision-making algorithms based on mathematical optimization for coordinated signaling control \cite{hunt1982scoot, luk1984two, gokulan2010distributed, ceylan2005genetic, zhang2018traffic}. However, those approaches are hard to deploy widely since they require computation-complex decision-making processes and ideal assumptions, which make them usually fall into local optima of decisions.}

Reinforcement Learning (RL) has recently shown excellent performance in a wide spectrum of practical scenarios, such as autonomous driving \cite{guan2022integrated}, Unmanned Aerial Vehicle (UAV) navigation \cite{ye2022multi}, and video games \cite{han2019grid}. These advances motivate researchers to develop RL-based policies for adaptive traffic signal control \cite{wei2021recent}. RL requires the formulation of a Markov Decision Process (MDP) for sequential decision-making tasks and fits parametric models to map the relationship between environmental observations and actions. By characterizing the models with Deep Neural Networks (DNNs), Deep Reinforcement Learning (DRL) with enhanced fitting performance can be used to solve complex decision-making problems. Nevertheless, {when both the state space and action space scale up, the learning capabilities of DRL agents will be limited due to the curse of dimensionality.} MARL is regarded as a potential alternative for networked system control problems since it learns coordinated control policies from interacting dynamics among agents \cite{tan1993multi}. {In this work, we aim to develop a new MARL algorithm for decentralized traffic signal control.}

{Considering the learning capacity of MARL in traffic signal control refers to the abilities of agents to acquire decision-making skills for making optimal signal control actions.} The problem of how to improve the learning capacity has been studied under the assumptions of a global reward and limited communications,
e.g., \cite{wiering2000multi, el2013multiagent, chu2019multi, zhou2020drle, wang2020stmarl}.
Specifically, the modular Q-learning principle in MARL is applied in \cite{wiering2000multi, el2013multiagent} to formulate traffic signal control policies. Their common goal is to capture the spatial correlation among traffic lights through a decomposable Q-function. In \cite{wiering2000multi}, a global goal is optimized by sharing global state information among agents while El-Tantawy et. al\cite{el2013multiagent} focus more on neighboring coordination to facilitate each agent in learning a joint control policy with its neighboring agents.
\begin{figure*}[!t]
\centering
\includegraphics[width=5.5in]{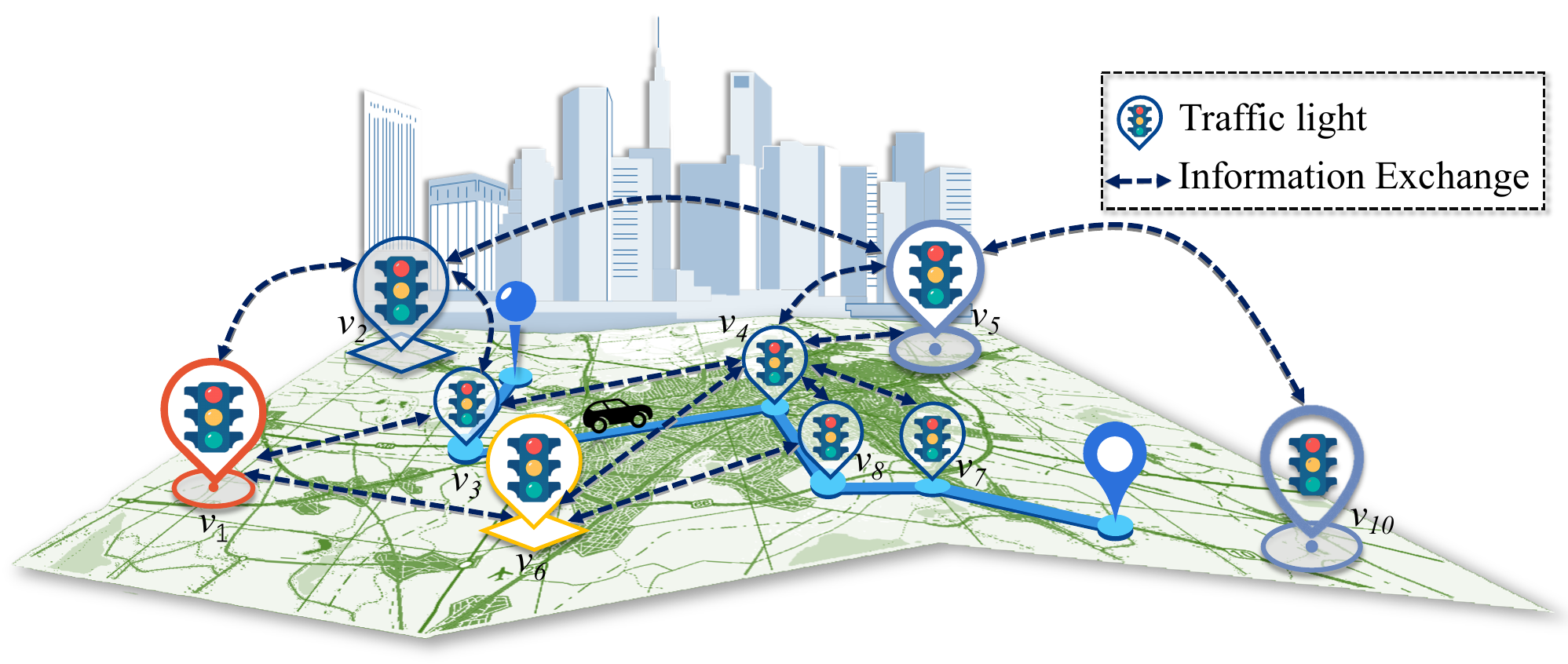}
\caption{Illustration of a decentralized traffic signal control system. MARL is used to deal with the challenges stemming from partial observation and non-stationary while efficiently capturing spatial-temporal correlations. 
}
\label{systemfigure}
\end{figure*} 

{In MARL, information exchange plays a fundamental role in improving agent cooperation, which accordingly determines the learning capacity of agents.} Considering the time-varying characteristics of traffic flow, recent studies, e.g., \cite{chu2019multi, zhou2020drle, tan2019cooperative}, investigate different information aggregation strategies in MARL. 
Specifically, based on a Multi-Agent Advantage Actor-Critic (MA2C) framework, Chu et. al\cite{chu2019multi} propose a spatially discounted information aggregation method, which captures the diverse impact of agents with different distances using a discount factor. This method scales down information of the agents with less effect. To aggregate information in a hierarchical way, Tan et. al\cite{tan2019cooperative} divide agents into different roles, i.e., a centralized agent and multiple regional agents. Each regional agent updates a local control policy based on regional observations while the central agent collects information from those regional agents and subsequently generates global actions. Similarly, a hierarchical information aggregation architecture is investigated in \cite{zhou2020drle} for an edge-assisted Internet-of-Vehicles (IoV) system. After evaluating the impact of vehicular communications on information aggregation quality, a theoretical optima proof is provided in \cite{zhou2020drle} for the proposed MARL algorithm.
Those information aggregation methods focus on the impact of spatial correlation on the learning capacity of agents. {Though promising, they ignore a pivotal property that the importance of spatial correlation among agents is not invariant over time. Therefore, the problem of how to capture the spatial correlation among other agents remains unexplored.}

In this work, we consider the decentralized control structure for traffic signal control, as shown in Fig.~\ref{systemfigure}. Decentralized control is suitable for multi-agent networked control systems since it allows individual learning of agents and alleviates the impact of interacting dynamics by executing coordinated behaviors of agents. {When formulating the decentralized control architecture, we focus on how to augment information aggregation on each individual agent and enable the learned policies to capture the spatial and temporal correlation from other agents in an effective way.}

{Notably, the underlying topology of traffic lights provides a directed graph that reflects the interacting relationships among agents.} The traffic flow transverses over the graph, which constitutes unstructured data that determines the spatial correlation among traffic lights. In addition, the spatial correlation among agents is naturally dynamic, i.e., the importance of neighboring agents for the current agent is time-varying. Capturing the diverse and time-varying importance of agents is conducive to achieving adaptive traffic signal control. Therefore, we propose a graph-based strategy to accomplish topology-aware information aggregation from unstructured data by developing a \emph{Stochastic Aggregation Graph Neural Network} (SA-GNN).
 

 




After aggregating information and constructing graph signals, the second problem is to capture the temporal correlation of traffic flow at individual agents and then learn decentralized control policies constrained by coordinated agent behaviors. We resort to a MARL framework and develop a diffusion convolution module based on Gated Recurrent Units (GRU) \cite{chung2014empirical}. Therefore, a new MARL framework that achieves graph learning on aggregated information for decentralized control systems is formed. Particularly, SA-GNN supports information sharing among agents in an asynchronous way. As such, the learned policy at each agent is invariant to system sizes so that system scalability is guaranteed. In essence, traffic signal control is taken as a decentralized and asynchronous decision-making problem in the new MARL framework. 

The contributions of this work include three-fold:

\begin{itemize}
\item{}
To achieve adaptive traffic signal control, we design a new decentralized control architecture that captures {the spatial and temporal correlation among agents with interacting dynamics caused by traffic flow.} As far as we know, this is the first work that exploits diffusion convolution in MARL for the networked system control problem. We shed light on the excellent learning capacity of Multi-Agent Graph Reinforcement Learning in extracting correlation-related information and capturing the stochastic correlation from unstructured signals.  

\item{}
Technically, we develop a topology-aware information aggregation strategy by characterizing the stochastic correlation among agents as a diffusion process, which provides a recursive way to effectively improve the information acquisition of agents with limited communications. 
{Under the framework of Multi-Agent Graph Reinforcement Learning, a new MARL algorithm is formulated.} It not only achieves adaptive feature learning but also makes the learned policies scalable to traffic systems with different scales. 
\item{}
We conduct extensive experiments to evaluate the performance of the proposed MARL algorithm. The evaluation results based on both real-world and synthetic traffic datasets verify that it improves {the learning capacity of agents, control efficiency of traffic systems, and travel experiences of drivers compared with state-of-the-art algorithms.}



 

\end{itemize}

\section{Related Works And Motivation}
In this section, we introduce the related works on learning-based traffic signal control and then illustrate the motivation of our work. 

For traffic signal control problems, the traditional optimization-based approach relies on many ideal assumptions to make optimization problems tractable \cite{wei2021recent}. In contrast, the RL-based approach avoids the deviation of assumptions from the real world. This is because RL, especially DRL, is able to learn the state abstraction and policy approximation from observed high-dimensional environmental data directly. The advances of DRL motivate researchers to carry out a series of studies to solve traffic signal control problems. 
 
\subsection{DRL  \& MARL for Traffic Signal Control} 
A straightforward way to apply DRL in traffic signal control problems is to combine global information in a centralized way and then make control decisions for agents. For example, a Deep Q Network (DQN) agent is developed in \cite{2018IntelliLight} for traffic signal control and evaluated in large-scale real-world datasets. In \cite{aslani2017adaptive}, authors compare the difference between Q-learning and Advantage Actor-Critic (A2C) and then develop an A2C-based control policy.
 
The centralized approach is not always feasible since both state and action space increase sharply with the rise of the system scale. {MARL distributes the decision-making process into agents, which makes practical implementation possible.} However, as environments are observed partially, it is usually difficult for each agent to reach a fast convergence. As such, learning the correlation among agents becomes the primary goal. To this end, the neighbors' information related to states or policies will be collected and encoded at each individual agent to enrich the local information space. Following this principle, 
we next introduce several popular works about MARL-based traffic signal control.
 
\subsection{MARL Based on Q-Learning and A2C}
The stochastic control process at a single traffic light with only local information can be characterized by an MDP. Instantaneous information exchange among agents is a key knob that makes MARL outperform single-agent RL. A fundamental problem in MARL is which kind of information should be shared to maximize the cooperation efficiency of agents with the lowest costs, which is first investigated in \cite{tan1993multi}.
Coordinated Q-learning (CQL) \cite{kok2006collaborative}\cite{guestrin2001multiagent} is an early attempt to investigate the problem by formulating a coordinated MDP in MARL. CQL requires decomposable Q-function $Q(s_t,a_t)=\sum_{i=1}^NQ_i(s_{t},a_{t})$ and iterative message passing $Q_i(s_{t},a_{t})\approx Q_i(s_{t},a_{i,t})+\sum_{j\in{\mathcal N_i}}\mathcal M_j(s_{t},a_{j,t},a_{\mathcal N_{j},t})$, where $N$ refers to number of agents, $\mathcal M_j(\cdot)$ denotes the message passed from neighbor $j$, and ${\mathcal N_i}$ denotes the neighbor set of agent $i$. $s_{t}$ represents the system state at time $t$, while $a_{t}$ and $a_{i,t}$ denote the global actions and action of agent $i$ at time $t$, respectively. The global state and reward are assumed to be accessible to each agent.
 
Compared to CQL, Independent Q-Learning (IQL)\cite{tan1993multi}\cite{tesauro2003extending} is a simplified and scalable approach since it omits the message passing in the Q function, denoted as $Q_i(s_t,a_t)=Q_i(s_t, a_{i,t})$. In essence, it takes the control policies of other agents as environmental observations to extend the local information space. During local learning, the information related to the actions of neighbor agents will be sampled. To mitigate the impact of outdated information from other agents, the temporal difference between the sampling time and updating time is estimated and then used to rectify the training process \cite{foerster2017stabilising}. 
 
MARL also benefits from the policy gradient method in learning multi-agent cooperation in traffic signal control systems. By replacing Q-learning in IQL with A2C, Independent A2C (IA2C) is formed in \cite{chu2019multi}. Similar to IQL, IA2C combines the policies of other agents to update the local return and incorporates other agents' states into the local state. As such, it suffers from the same problems as IQL. 
To address that, MA2C\cite{chu2019multi} uses a factor to discount the information related to policies and states of neighbors with different distances. By combining local state and discounted information from neighbors, the parameterized control policy at agent $i$ could be denoted as:
\begin{equation}
    \pi_{i,t}=\pi_{\theta_i}(\cdot|s_{i,t},s_{\mathcal N_i,t},\pi_{\mathcal N_i,t-1}),
\end{equation}
where $\theta$ parameterizes the policy.
Moreover, MA2C transforms the global return into a discounted global return, in order to release the complexity of global cooperation. Let $r_{t,j}$ denote the reward of neighbor agent $j$, and the local reward at agent $i$ after combination is denoted as:
\begin{equation}
\label{comreward}
    r^{com}_{i,t}=\sum_{d=0}^{D_i}(\sum_{j\in{\mathcal N_i|d(i,j)=d}}\alpha^d{r_{j,t}}),
\end{equation}
where $D_i$ is the maximum distance. $d(i,j)$ denotes the distance between agent $j$ and $i$. The second summation operation is limited by the term $j\in{\mathcal N_i|d(i,j)=d}$, which denotes that agent $j$ belongs to the neighbor set ${\mathcal N_i}$ and locates at a distance of $d$ from agent $i$.

{In the works of MARL-based traffic signal control}, the state is defined to reflect the property of traffic signal control systems, such as dynamics. The most commonly used items include \emph{real-time signal phases of traffic lights}, \emph{number of vehicles waiting in the lanes of traffic lights}, \emph{cumulative delay of vehicles on their trip} \cite{wei2021recent}\cite{chu2019multi}. In terms of action definition, the elements related to signal timing are popular, such as \emph{signal phase} \cite{chu2019multi}, \emph{signal phase switch} \cite{el2013multiagent}, \emph{signal phase duration} \cite{aslani2017adaptive}.

\subsection{LSTM Empowered MARL} 
In addition to spatial correlation, temporal correlation is also an important property in traffic signal control systems. Fortunately, long-term-short-memory (LSTM) is verified as an effective neural network architecture in learning short-history memory \cite{1997Long}. Both MA2C and Spatio-Temporal MARL (STMARL) \cite{wang2020stmarl} incorporate LSTM into MARL when developing traffic signal control policies. Specifically, STMARL is similar to IQL because it incorporates the observation information of neighbor agents into the local Q-function. It uses a graph attention mechanism to learn the spatial correlation among agents and therefore it outperforms IQL. More spatially related information could be captured by implementing higher-order relation reasoning. Among them, LSTM is adopted to capture the temporal correlation of historical traffic information at each agent. As such, each agent learns its control policy independently based on the local state that combines both spatial and temporal information. However, STMARL inevitably suffers from a huge computation burden when capturing more spatial information in the partially observed environment.
 
 
\subsection{Motivation}

According to recent works, traffic signal control based on MARL is challenged by two practical issues, i.e., partial observability and non-stationary MDP.
In addition, MARL inherently suffers from low convergence performance and requires extra communication overhead. Motivated by that, this work
provides an information aggregation strategy and then develops a graph learning module in MARL, in order to ensure scalability and simultaneously improve the learning capacity. Graph learning has been used in traffic signal control systems to develop control policies, such as STMARL \cite{wang2020stmarl}. However, the STMARL-empowered policy has no consideration of the stochastic correlation among agents. Besides, in traffic signal control problems,
the scale of information collected from neighbors should be controlled while the effectiveness of information must be guaranteed, which significantly challenges traditional GNN. 

\section{Problem Formulation}
{In this section, we first present the system overview and then formally define the problem of decentralized traffic signal control.}




\subsection{System Overview}

{We consider a system containing $N$ traffic signal lights and each of them is equipped with a $compute~unit$ to update the local control policy.} 

{The connections among traffic lights form an underlying directed graph, which can be characterized by $G=\{\mathcal{V}, \mathcal{E}\}$.} $\mathcal{V}$ denotes the set of traffic lights, and two traffic lights $v_i$ and $v_j$ are connected via edge $e_{i,j}\in\mathcal{E}$. For brevity, in the following, $v_i$ and $v_j$ are simplified as $i$ and $j$, respectively, while ${ij}$ is used to replace $e_{i,j}$. We further characterize the system as a multi-agent networked control system, where each traffic light (a.k.a agent hereafter) observes the environment, executes actions based on its learned policy, and communicates with immediate neighboring agents. 
The state of agent $i$ at time $t$ is denoted as $s_{i,t}\in{\mathcal S}$ while the corresponding action of signal adjustment is $a_{i,t}\in{\mathcal A}$. Both $s_{i,t}$ and $a_{i,t}$ are discrete scalars. Although traffic flow in the system varies in an arbitrary way, the actions of an individual agent will certainly affect the traffic flow observed at nearby agents, resulting in spatial correlation of actions among agents. 
{In order to capture this property, we group local states and actions into the global system observation vector
$\mathbf{O}(t)=[\mathbf{s}(t),\mathbf{a}(t)]$, where $\mathbf{s}(t)=[{s}_{{1},t},{s}_{2,t},...,{s}_{N,t}]$ and $\mathbf{a}(t)=[{a}_{1,t},{a}_{2,t},...,{a}_{N,t}]$. 
}





Traditionally, the networked system is controlled in a centralized way, i.e., finding the optimal control actions for all agents at a centralized server. This pattern highly relies on the information collected globally, which, however, inevitably degrades the efficiency of signal control when the system is scaled up. Decentralized control allows agents to execute independent or coordinated actions based on local policies with information exchange with neighbors.
We focus on the decentralized way and aim to balance the scalability and optimality of traffic signal control. Information exchange plays a fundamental role in decentralized control since it helps capture spatial correlation among agents, which motivates us to first explore an efficient information aggregation strategy.

In decentralized settings, we characterize system dynamics in a discrete approach, i.e., the system state is recorded at each discrete time instance $t$ while control actions are taken at each sampling time $T_s$. 
To aggregate more useful information, we define those nodes who reach agent $i$ via $k$ hops as the $k$-hop neighbors of agent $i$, denoted as ${\mathcal N_i^k}\subseteq\mathcal{V}$. When $k=1$, the immediate ($1$-hop) neighbors of agent $i$ are denoted as ${\mathcal N_i}=\{j|ij\in \mathcal{E}\}$ by neglecting the index of $k=1$. 
{The term \emph{hop} refers to the minimum number of edges connecting two agents}. $k$-hop means the path on which vehicles travel includes $k$ essential traffic lights. The $k$-hop neighbors of agent $i$ at time $t$ are the ($k-1$)-hop neighbors of the agents from the $1$-hop neighbors of agent $i$ at time $t-1$, i.e., ${\mathcal N_{i,t}^k}=\{j'\in\mathcal N_{j(t-1)}^{{k-1}}|j\in{\mathcal N_{i,t}}\}$.
$\mathcal{N}_{j,{t-1}}^{k-1}$ refers to the set of active neighbors within ${\mathcal N_{j}^{k-1}}$ that cause non-negligible influence on the traffic flow of traffic light $j$ at time $t-1$. ${\mathcal N_{i,t}}$ refers to the $1$-hop active neighbors of agent $i$ at time $t$.
With the multi-hop neighbor definition, the locally available information of each agent thus involves delayed neighboring information. We define the locally available information of agent $i$ at time $t$ as: 
\begin{equation}
\label{informationhistory}
    \mathcal{O}_{i,t}=\bigcup\limits_{k=0}^{K-1}\{ \mathbf{O}_j(t-k+1), {s}_{j,(t-k)}, {a}_{j,(t-k)}  |j\in{\mathcal N_{i,t}^k} \},
\end{equation}
{where $K$ delineates the upper limit of spatial-temporal correlation.}
{To minimize communication overhead, limited communications are adopted in our work,} i.e., information exchange occurs only between two immediately neighboring agents. We assume that each agent saves the initial global
observation information $\mathbf{O}(0)$. 
At time $t$, each agent individually combines both local information and immediate neighboring information, resulting in the formation of $\mathbf{O}_i(t)$ at agent $i$. That is, $\mathbf{O}_i(t)$ is obtained by combining ${s}_{i,t}$ and ${a}_{i,t}$ as well as \{$\mathbf{O}_j(t-1), {s}_{j,t-1}, {a}_{j,t-1}|j\in{\mathcal N_{i,t}^k}$\} into $\mathbf{O}_i(t-1)$. 
{Consequently, the local information $\mathbf{O}_i(t)$ is updated in a recursive way. We use a specific symbol $\mathcal{O}_{i,t}$ to represent the combination of all the information locally available at agent $i$. Since $\mathbf{O}_i(t)$ is inherent information of agent $i$,} we omit it and finalize the form in ($\ref{informationhistory}$).
Notably, with recursive combination and limited communications, the construction of $\mathbf{O}_i(t)$ reduces communication overhead compared with other methods that require real-time global information exchange.

\subsection{Problem Formulation}
\label{problemformulation}
{The decentralized control problem lies in finding the optimal policies that minimize long-term costs restricted to the information structure in (\ref{informationhistory}). It is famously difficult except in some particularly simple scenarios \cite{witsenhausen1968counterexample, eksin2014bayesian}.} The problems with information structure (\ref{informationhistory}) also challenge traditional heuristic methods according to \cite{witsenhausen1968counterexample}.  
This is because each agent makes local decisions based on its limited observations, while the decisions must be conducive to a collective task. {To develop learned heuristics for optimal policies, we next provide a formal description of the decentralized control problem.}

Specifically, the dynamics of environment states in traffic signal control systems is Markovian.
{
The traffic signal control problem can be formulated as an MDP, denoted as $<\mathcal S, \mathcal A, P, R, \gamma>$. Specifically, the state at each agent is defined as $s_{i,t}=\{{\rm wave}_t[g]\}_{g\in{\mathcal{G}_{ji}},{ji}\in\mathcal{E}}$, where ${\rm wave}_t[g]$ is reflected by the number of vehicles at each incoming lane $g$ of the agent. ${\mathcal{G}_{ji}}$ refers to the set of lanes on edge $ji$. The action is selected from the available signal phases of traffic lights by following \cite{prashanth2010reinforcement}. After observing state $s_{i,t}\in\mathcal S$, the agent decides on an appropriate action $a_{i,t}\in{\mathcal A}$ for controlling its attached traffic light.}
The state then transfers to the next state $s_{i,t+1}$ with the transition dynamics characterized by a transition distribution $P\{s_{i,t+1}|s_{i,t}, a_{i,t}\}$. 
{With the goal of maximizing long-term return discounted by $\gamma$, the total return of agent $i$ is defined as $R_{i,t}=\sum_{\tau=t}^{+\infty}\gamma^{\tau-t}r_{i,\tau}$, where $r_{i,\tau}$ is the step reward received by agent $i$.} {$r_{i,\tau}$ is calculated as a combination of $wait$ and $queue$. For the incoming lanes at traffic light $i$, ${\rm wait}_{i,\tau}$ is the maximum cumulative delay of the first vehicles present in those lanes while ${\rm queue}_{i,\tau}$ refers to the queue length denoted by the number of vehicles along each incoming lane. Both of them are calculated in a fixed duration $\Delta{\tau}$.

{
Formally, $r_{i,\tau}$ is defined as 
\begin{equation}
r_{i,\tau}={\rm queue}_{i,\tau}+\omega {\rm wait}_{i,\tau},  
\end{equation}}
where $\omega$ is a coefficient to balance the traffic congestion and trip delay. ${\rm queue}_{i,\tau}$ is defined as: 
{
\begin{equation}
\label{queueperagent}
{\rm{queue}}_{i,\tau} = \sum_{{g}\in{\mathcal{G}_{ji,{ji}\in\mathcal{E}}}} {\rm{veh.}}_{\tau+\Delta{\tau}}[g].
\end{equation}}
}

From the global perspective, the traffic signal control problem is a cooperative game in which all agents participate to optimize a global reward. {We thus define the global reward utility as
\begin{equation}
\label{globalutility}
    u(R)=\lim_{T\rightarrow+\infty}\frac{1}{T}\sum_{\tau=0}^{T}\sum_{i\in\mathcal V}\gamma^{\tau}r_{i,\tau}.
\end{equation}
Denote by $p_i(\mathcal{O}_{i,t})$ the local policy of agent $i$ at time $t$, the joint policy is defined as a vector $\mathbf{P}(\mathcal{O}_{t})=[p_1(\mathcal{O}_{1,t}),p_2(\mathcal{O}_{2,t}),...,p_N(\mathcal{O}_{N,t})]$. }

{Although MARL provides a suitable framework for decentralized control, it is still challenging since there is no explicit knowledge to accurately characterize system transition dynamics and the relationship between reward and actions \cite{witsenhausen1968counterexample, tolstaya2020learning, eisen2019learning, gama2018convolutional}.} 
{We use a parameterized policy $\pi_{\theta_i}(\mathcal O_{i,t})$ to substitute the local control policy $p_i(\mathcal{O}_{i,t})$, where ${\theta_i}$ is the parameter tensor. Hence, the action vector is defined as $\mathbf{a}(t) \triangleq [\pi_{\theta_i}(\mathcal O_{i,t}, \mathbf{O}_i(t-1), s_{i,t}), i=1...N ]$.
The decentralized control problem is thus formalized as 
\begin{equation}
\begin{aligned} \label{maximationproblem}
&\max_{\{\mathbf \pi_{\theta_i}(\cdot), i=1...N\}} \quad u(R)\\
&\begin{array}{r@{\quad}r@{}l@{\quad}l}
\end{array}
\end{aligned}
\end{equation}
}
The problem formulation above has two advantages. First, the dimension of ${\theta_i}$ is controllable and independent of the number of agents, which means that the learned policy can be applied to systems with different scales. 
Second, the tensor that parameterizes the local policy could be shared with other agents, which avoids learning multiple diverse policies for agents. 
Nevertheless, a new challenge arises since the historical information dimension varies across agents because the number of adjacent neighbors for individual agents is diverse. 

{The notations and their descriptions used in this work are shown in TABLE \ref{notations}.}

\begin{table}
\centering
  \begin{threeparttable}
    \caption{{Key Notations in System Overview}}
     \begin{tabular}{lllll}
        \toprule
        Notations & Descriptions \\
        \midrule
        $\mathcal{V}$    & Set of traffic lights (agents)     \\
        $\mathcal{E}$    & Set of edges     \\
        $v_i$  & Traffic light $i$ (agent $i$)     \\
        $e_{i,j}$  & Edge connecting $v_i$ and $v_j$     \\
        $\mathcal S$ & State space \\
        $\mathcal A$ & Action space \\  $\mathbf{a}(t)$ & Vector involves all actions of agents  \\      
        $\mathbf{s}(t)$ & Vector involves all states of agents \\
        ${a}_{i,t}$ & Action determined by agent $i$ at time $t$ \\
        ${s}_{i,t}$ & State observed by agent $i$ at time $t$\\
        ${\mathcal N_i}$ & Set of immediate neighbors of agent $i$ \\
        $\mathbf {O}(0)$ & Initial global observation information \\
        $\mathbf {O}_i(t)$ & Global information individually updated by agent $i$ \\
        $P$ & State transition probability \\
        $r_{i,\tau}$ & reward received by agent $i$ at $t$ \\
        $R_{i,t}$ & Discounted long-term return obtained by agent $i$ \\
        $\gamma$ & Discounted factor \\      $p_i(\mathcal{O}_{i,t})$ & Local policy of agent $i$ \\
        $\mathbf{P}$ & Joint policy set \\
        $\pi_{\theta_i}(\mathcal O_{i,n})$ &  Parameterized policy \\
        $K$ & Maximum number of hops  \\   
        ${\rm w}_{i,j}$ & Affinity weight from $v_i$ to $v_j$ \\
        $\bold{W}$ & Weight matrix of edges \\
        $\alpha$ & Restart probability of random walk \\
        $\mathbf{D}_{\mathbf{W}}^{out}$ & Diagonal matrix of out degrees \\
        $\mathcal{P}_{v_i}$ & Graph shift operator (GSO) \\
        $\bm{\mathcal{P}}$ & Diffusion matrix \\
        $\mathbf y^{Agg}_{i,t}$ & Traffic aggregation sequence \\
        $L$ & Number of neural network layers \\
        $\mathcal{G}_{ji}$ & Set of incoming lanes on edge $ji$ \\
        $\Psi_l^{fg}$ & Filter tap \\
        $\mathbf x_l$ & Intermediate feature of layer $l$ \\
        $\mathbf z_l$ & Output feature of layer $l$ \\
        $\theta_i$ & Parameter of the policy network at agent $i$ \\
        $w_i$ & Parameter of critic network at agent $i$ \\
        $B$ & Batch in replay buffer \\
        $\mathfrak{R}_t$ & Reset gate \\
        $\mathfrak{Z}_t$ & Update gate \\
        $\mathbf h_t$ & Recurrent hidden unit \\
        $T$ & Time horizon of episode\\
        ${\rm veh.}$ & Simplified notation of \emph{vehicle} as index \\ 
        $\tau$, $t$, & Time variable \\
        \bottomrule
     \end{tabular}
     \label{notations}
  \end{threeparttable}
\end{table}

\section{SA-GNN for Traffic Light Control}


This section illustrates a new information aggregation strategy, called SA-GNN, to address the challenges identified above. With SA-GNN, each agent is able to aggregate information from neighbors with different distances and then output intermediate features for learning spatial correlation among agents. Limited communications are required by SA-GNN because information is aggregated in a recursive way. 

 

\subsection{Modelling Stochastic Correlation Based on Random Walk}
 
The correlation among agents is affected by many practical factors in a traffic signal control system, including current phases of signal light, dynamic traffic volumes, and speed limitation of vehicles. 
{Traditionally, the raw message reflecting correlation is directly included to form a correlation model and then develop control policies
\cite{yue2021root, wang2019network},} which, however, inevitably leads to intensive computation. Moreover, a scalar message is usually inadequate to capture the stochastic correlation in complex traffic environments. 
With recent advances in machine learning, Neural Network (NN) based models are adapted to extract correlation-related features in traffic scenarios, e.g., \cite{wang2020stmarl, wu2021dynstgat}. {The low-dimensional information extracted by trained NNs significantly improves the traffic signal control performance.} Nevertheless, the stochastic correlation among agents, which is naturally determined by the stochastic dynamics of traffic flow, has not been considered in the literature. 
 
In SA-GNN, to aggregate more neighboring information and then augment features related to stochastic correlation at each agent, {we extend the correlation to $K$-hop neighbors.
Note that the information aggregation procedure is independent of downstream tasks, accordingly forming a task-independent feature learning problem.}

We first model the stochastic correlation among agents by a diffusion process. 
Considering that connected agents establish an irregular topology over the underlying road network, we opt to characterize the diffusion process by a bidirectional random walk \cite{teng2016scalable}. 
{With a flexible definition, the random walk is able to reflect how traffic flow affects the stochastic correlation among agents on a given underlying road network.} Based on diffusion process modeling, an agent is able to combine high-order neighborhood information when learning features locally.
We incorporate a non-negative weight matrix $\bold{W}$ into the network graph and thus form a new graph definition $G=\{\mathcal{V}, \mathcal{E}, \bold{W}\}$. 

In our work, $\bold{W}$ is defined as a weighted adjacency matrix whose element ${\rm w_{i,j}}$ indicates the affinity weight from agent $i$ to agent $j$. {With $\bold{W}$, the out-degree diagonal matrix is $\mathbf{D}_{\mathbf{W}}^{out}$.} {The stationary distribution of random walk with a restart probability $\alpha$ at agent $i$ is calculated as:
\begin{equation}
\label{RWdistribution}  
\mathcal{P}_{i}={\alpha}\sum_{d=0}^{D }(1-\alpha)^d(\bold{W^T}(\mathbf{D}_{\mathbf{W}}^{out})^{-1})^d\mathbf{1}_{v_i},
\end{equation}
where $\mathbf{D}_{\mathbf{W}}^{out}=diag([d_1^{out},...d_n^{out}])$ denotes the diagonal matrix of out degrees.} The diffusion process thus can be characterized by two matrices, i.e., $\bold{W^T}(\mathbf{D}_{\mathbf{W}}^{out})^{-1}$ and $\bold{W}(\mathbf{D}_{\mathbf{W}}^{in})^{-1}$, which are respectively transition matrices and the reverse one. 
Considering global agents in $G$, we define a diffusion matrix as $\boldsymbol{\mathcal{P}}=({{\mathcal{P}}_{v_1}, {\mathcal{P}}_{v_2},...,{\mathcal{P}}_{v_n}})^T$ by combining random walk matrices $\bold{W^T}(\mathbf{D}_{\mathbf{W}}^{out})^{-1}$ in a convex way. 
Note that the sparsity of $\boldsymbol{\mathcal{P}}$ implicitly reflects the traffic conditions at each agent. {Here, $[{\boldsymbol{\mathcal{P}}}]_{i,j}$ can be interpreted as the probability of traffic flow visiting agent $j$ on a random walk starting from agent $i$.} $D$ is the maximum diffusion step as we set a $D$-step truncation of the diffusion process.

\subsection{Topology-Aware Information Aggregation}

 

In this part, we illustrate how to aggregate the neighboring information considering the stochastic relationships of agents. {Particularly, we focus on the capability of agents that capture time-varying impact due to the behaviors of neighbors.}


Recall that (\ref{informationhistory}) provides the definition of historical information, which indicates that agent $i$ collects information from neighbors with maximal $K$ hops. Information aggregation aims to rearrange the information collection and transform local information into the signals supported on graphs. 
The matrix in (\ref{RWdistribution}) can be alternatively interpreted as graph shift operator (GSO) for graph signals according to \cite{sandryhaila2014big, chen2015discrete}. {That is, $\boldsymbol{\mathcal{P}}$ is used to aggregate neighborhood information through multiplication operations with which graph signals could be developed.} We assume that $\boldsymbol{\mathcal{P}}_t$ is the updated stationary probabilities at aggregation time $t$ considering that $\bf{W}$ varies over time. {Let $\mathbf y^0_{t}=\mathbf {O}(t)$, i.e., the initial observation of agents at time $t$, the one-step information aggregation is denoted as
\begin{equation}
\label{step0_informationagg}
\mathbf y^1_{t}=\boldsymbol{\mathcal{P}}_{t}\mathbf {y}^0_{t-1}.
\end{equation}}
Note that $\mathbf y^1_{t}$ could be obtained at each agent by the information exchange with immediate neighbors. For example, the $i$-th element in $\mathbf y^1_{t}$ could be obtained at agent $i$ as: 
\begin{equation}
\label{step0_informationagg}
[\mathbf y^1_{t}]_i=\sum_{j\in{\mathcal N_{i}}}[\boldsymbol{\mathcal{P}}_{t}]_{i,j}[\mathbf {y}^0_{t-1}]_j,
\end{equation}
which is a linear and local operation since only neighboring information aggregation is carried out.
We further write the two-step information aggregation as:
\begin{equation}
\begin{aligned}
\label{step2_informationagg}
\mathbf y^2_{t+1}&=\boldsymbol{\mathcal{P}}_{t+1} \mathbf y^1_{ t}\\
&=\boldsymbol{\mathcal{P}}_{t+1} \boldsymbol{\mathcal{P}}_{t} \mathbf y^0_{t-1}.
\end{aligned}
\end{equation}
Following this line of recursive aggregation, we are able to obtain the $k$-step information aggregation respecting the $k$-hop information collection in (\ref{informationhistory}), denoted as 
\begin{equation}
\label{step3_informationagg}
\mathbf y^k_{t}=\boldsymbol{\mathcal{P}}_{t} \mathbf y^{k-1}_{t-1}=(\boldsymbol{\mathcal{P}}_{t}\boldsymbol{\mathcal{P}}_{t-1}...\boldsymbol{\mathcal{P}}_{t-k+1})\mathbf y^0_{t-k}
\end{equation}
By collecting the aggregated signals at each step, the local information sequence at agent $i$ is: 
\begin{equation}
\label{step4_informationagg}
\mathbf y^{Agg}_{i,t}=[[\mathbf y^0_{t}]_i, [\mathbf y^1_{t}]_i, [\mathbf y^2_{t}]_i,...,[\mathbf y^{K-1}_{t}]_i].
\end{equation}
The sequence in (\ref{step4_informationagg}) is also named as 
a traffic aggregation sequence. 

\subsection{Graph Convolution Structure}
\label{graphconvolutionstructure}

The traffic aggregation sequence defined in (\ref{step4_informationagg}) exhibits a time structure by transforming the information from irregular topology to structured graph signals. It can be used as the input to convolution operations. We thus provide the principle of graph convolution in SA-GNN.

{We first define the graph convolutional filter generalizing from the convolutional filter of Convolutional Neural Network (CNN).} For the convolution layer $l$ in a $L$-layer SA-GNN, the output feature is denoted as $\mathbf z_l$, which will be taken as the input of the graph filter performed locally to produce the features of layer $l+1$, i.e., $\mathbf z_{l+1}$. The initial input into $1$-th layer is the traffic aggregation sequence $\mathbf y^{Agg}_{i,t}$.
Denote as $\Psi_l$ the $K$-tap filter that contains $K-1$ filter taps, which is used to calculate the intermediate feature of layer $l$ based on the output of layer $l-1$: 
\begin{equation}
\mathbf x_l = \Psi_l \mathbf z_{l-1}=\sum_{k=0}^{K-1}\psi_{lk} \mathbf z_{l-1}.
\end{equation}
The feature output of layer $l$ is calculated as  
\begin{equation}
\mathbf z_l = \phi_l(\mathbf x_{l}),
\end{equation}
where $\phi_l$ is a pointwise nonlinearity function. 

{Notably, frequent calculation of $\boldsymbol{\mathcal{P}}_{t}$ is necessary to accomplish the graph convolution above, which will definitely lead to redundant computation, especially in large-scale traffic systems.} We thus consider graph convolution with the fixed graph shift operator, i.e., the graph shift operator $\boldsymbol{\mathcal{P}}$ is constant. Recall that $\bold{W^T}(\mathbf{D}_{\mathbf{W}}^{out})^{-1}$ is the transition matrix, we further adopt the reverse transition matrix $\bold{W}(\mathbf{D}_{\mathbf{W}}^{in})^{-1}$ to  
develop the graph convolution operation as
\begin{equation}
\label{fixedaggregation}
\mathbf y^k_{t}=\sum_{j=0}^{k-1}( 
(\bold{W^T}(\mathbf{D}_{\mathbf{W}}^{out})^{-1})^j + (\bold{W}(\mathbf{D}_{\mathbf{W}}^{in})^{-1})^j
)\mathbf y^0_{t-k}.
\end{equation}
With (\ref{fixedaggregation}), the $K$-hop traffic aggregation sequence in the fixed graph can be constructed. To cater for the extension of graph shift operation in (\ref{fixedaggregation}), each $\psi_{lk}$ is extended into $\psi_{lk,1}$ and $\psi_{lk,2}$, corresponding to the transition matrix and its reverse matrix. The graph convolution with graph signal $\mathbf y^k_{n}$ will reduce the computation overhead when aggregating global information. 

We further augment the expressive power of the graph operation by setting that each layer is able to produce multiple features. To accomplish that, each layer will be processed by a bank of filter taps, denoted as $\Psi^{fg}_l$. That is, it is assumed that layer $l-1$ outputs $F_{l-1}$ features, they will be the input of layer $l$ and each of them is processed by a $F_{l}$ filter taps $\Psi^{fg}_l$. Applying the operation above to each input feature,
layer $l$ finally produces $F_{l-1} \times F_{l}$ intermediate features, i.e., 
\begin{equation}
\label{graphintermediate}
\mathbf x_l^{fg} = \Psi_l^{fg} \mathbf z_{l-1}^{f}=\sum_{k=0}^{K_l-1}\psi_{lk}^{fg} \mathbf z_{l-1}^{f}.
\end{equation}
The output of layer $l$ is then calculated as: 
\begin{equation}
\label{graphconvolutionmultiple}
\mathbf z_{l} = \phi_l(\mathbf x_{l}^{fg})=\phi_l\sum_{f=1}^{F_l}\mathbf x_{l}^{fg}=\phi_l\sum_{f=1}^{F_l}\Psi_l^{fg} \mathbf z_{l-1}^{f},
\end{equation}
where $\mathbf z_{l}$ has $F_l$ features.
 

The process above indicates the implementation of graph convolution on graph signals, where $[\Psi^{fg}_1,...\Psi^{fg}_L]$ include the set of learnable parameters to complete linear transformation for multiple features at each layer. Therefore, a $L$-layer SA-GNN architecture could be formed by using $L$ times recursive graph convolution operation. 

\begin{figure*}[!t]
\centering
\hspace{-2.4cm}
\begin{minipage}[t]{0.4\textwidth}
\centering
\includegraphics[width=3in]{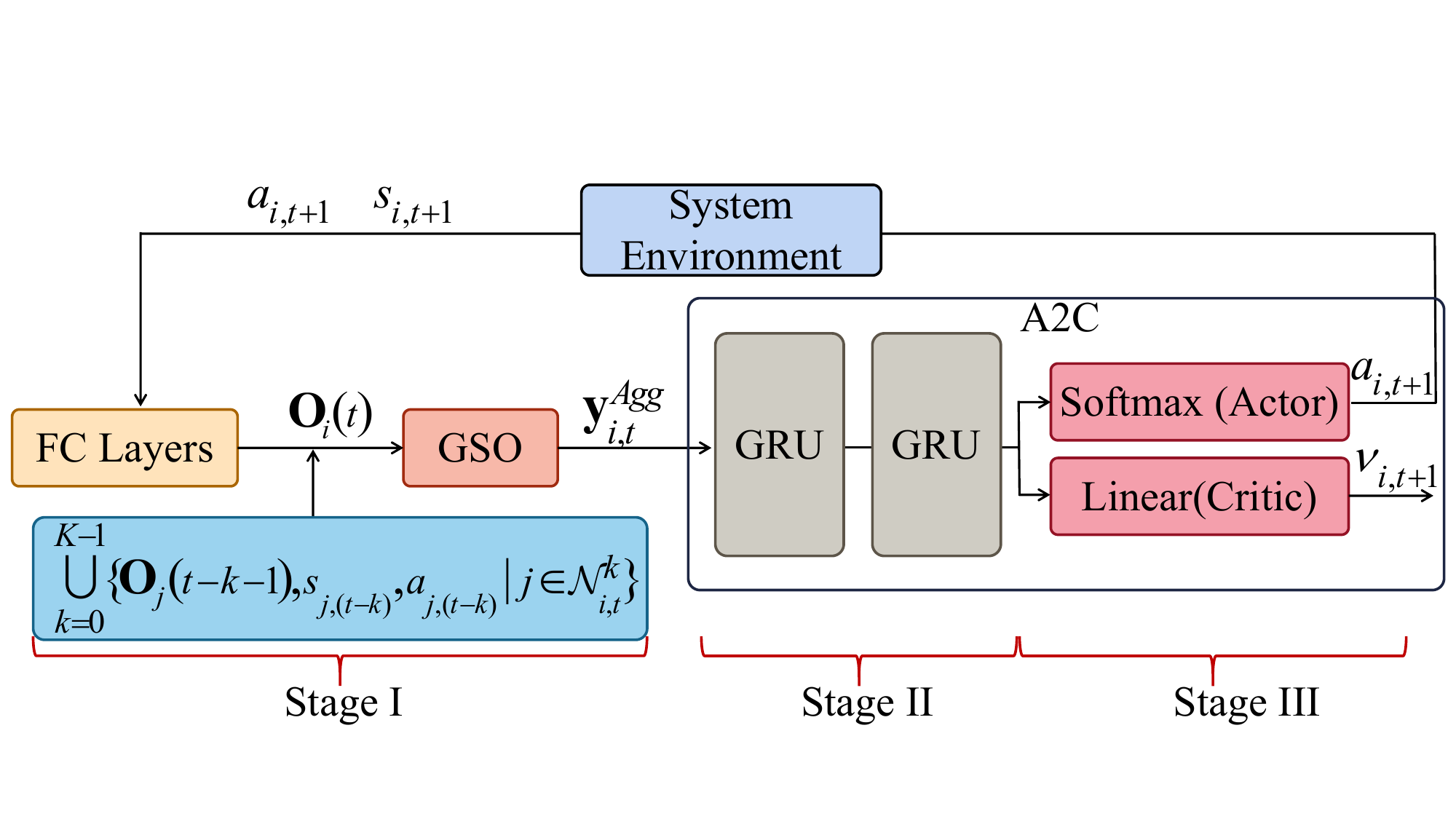}
\caption{Block diagram of algorithm pipeline at agent $i$. 
}
\label{BlockDiagram}
\end{minipage}
\hspace{0.6cm}
\begin{minipage}[t]{0.42\textwidth}
\centering
\includegraphics[width=3.9in]{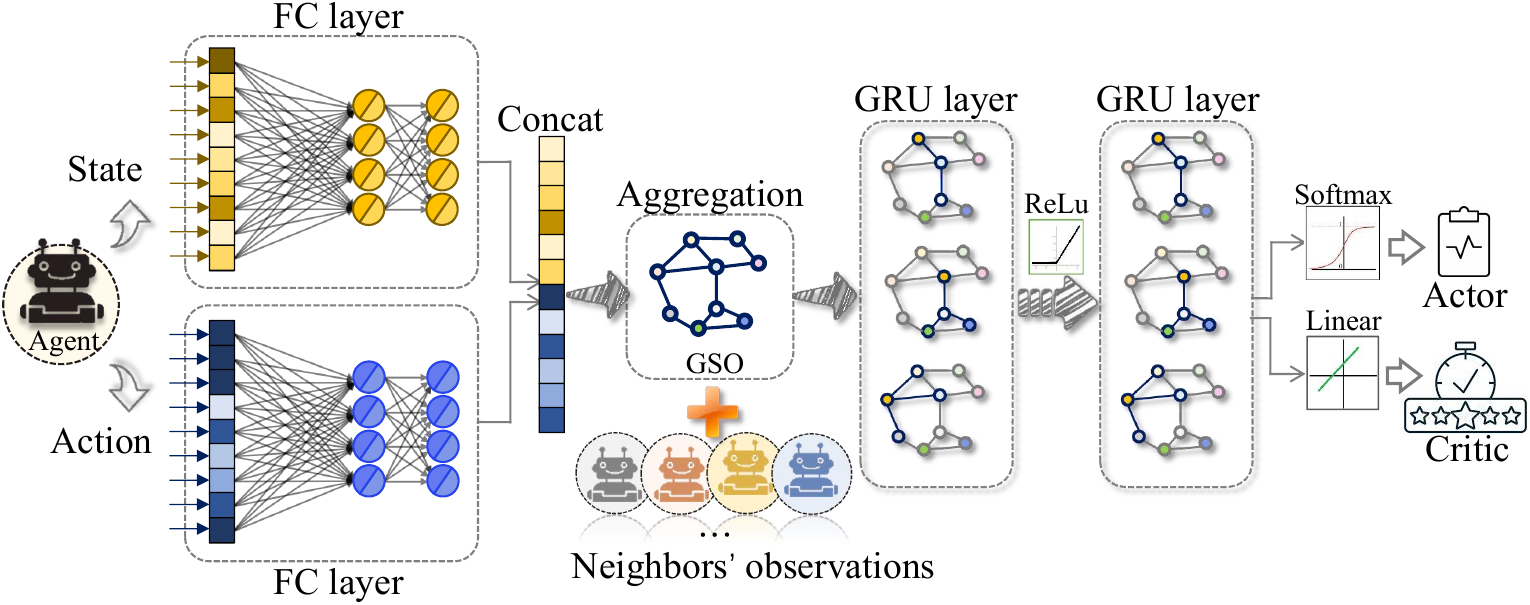}
\caption{Algorithm architecture.}
\label{NetworkArchitecture}
\end{minipage}
\end{figure*} 

\section{Diffusion Convolution Empowered MARL}


SA-GNN provides a scalable approach for agents to aggregate global information. {Under the structure of graph convolution structure defined above, we then exploit the multi-agent graph learning framework for the decentralized control problem in (\ref{maximationproblem}).}
 

{Specifically, SA-GNN extends the local information space of each agent by a recursive aggregation way.} Note that SA-GNN is specialized in spatial correlation. {To further augment the temporal correlation, we develop a GRU-based graph convolution module and then incorporate it into the MARL framework.} With SA-GNN and GRU-based graph convolution, each agent in MARL is able to capture the spatial and temporal correlation of neighbors.

 
 

\subsection{{Formulation of Multi-Agent Graph Reinforcement Learning}}


{The problem defined in (\ref{maximationproblem}) is for a multi-agent networked system where each agent participates in a collaborative game to optimize the global traffic flow.} Immediate information exchange between two agents is performed via the connections over the corresponding edge. Considering the procedure of information aggregation in $G=\{\mathcal{V}, \mathcal{E}, \bold{W} \}$, we further characterize the MDP of the multi-agent networked system as $<G, \{ \mathcal S_i, \mathcal A_i\}_{i\in\mathcal{V}}, P, R,>$, where $\mathcal S_i$ and $\mathcal A_i$ respectively denote the state space and action space at agent $i$. {The definitions of both state and action are similar to that in \ref{problemformulation}. $s_{i,t}$ is reflected by the number of vehicles at each incoming lane within $50$m of the intersection.} Based on the local policy, each agent performs action $a_{i,t}$ at each time $t$ after observing state $s_{i,t}$. With information aggregation, the local control policy at agent $i$ is constructed by encoding locally available information:
\begin{equation}
    \pi_{i}=\pi_{\theta_i}(\cdot|f(\mathcal O_{i,t}, \mathbf{O}(t-1)_i, s_{i,t})).
\end{equation}
Thus, the action at time $t$ is obtained as 
$a_{i,t}\sim\pi_{i}$. A step reward at agent $i$ is observed. For decentralized control, we assume that the transitions at an individual agent are independent of each other. That is, the local transitions are obtained as
\begin{equation}
\begin{split}
\label{localtra comrewardnsition}
    &P_i(s_{i,t+1}|s_{i,t}, a_{i,t}) = \\ &\sum_{\{a_{j,t},j\in\mathcal N_i^k\}} \prod_{j\in\mathcal N_i^k}\pi_j(a_{j,t}|\cdot) P(s_{i,t+1}|s_{i,t}, a_{i,t}, a_{j,t}).
\end{split}
\end{equation}
The global reward utility $u(R)$ is then 
calculated according to (\ref{globalutility}).

 

{We use A2C as the basic learning framework of MARL. By managing the $action$-$value$ function and $state$-$value$ function simultaneously, A2C has been demonstrated as an efficient RL framework.} 
To define decentralized actor-critics, we characterize the optimal policy at each agent by the parametric model $\{\pi_{\theta_i}\}_{i\in{\mathcal{V}}}$. The value function at each agent is characterized by the parametric model $\{V_{w_i}\}_{i\in{\mathcal{V}}}$. 

The loss function used to update the actor is defined as 
\begin{equation}
\begin{split}
\label{policyloss}
\mathcal{L}(\theta_i)=&-\frac{1}{|B|}\sum_{\tau\in{B}}
(
{\rm log}{\pi}_{\theta_i}(a_{i,\tau}|s_{i,\tau},a_{\mathcal{N}_i,\tau-1})\hat{A}_{i,\tau} -\\ &\beta\sum_{a_i\in{\mathcal{A}_i}}\pi_{\theta_i}{(a_i|s_{i,\tau}){\rm log}\pi_{\theta_i}(a_{i,\tau}|s_{i,\tau},a_{\mathcal{N}_i,\tau-1})}
),
\end{split}
\end{equation}
where $\hat{A}_{i,\tau}=\hat{R}_{i,\tau}-V_{w_i^-}(s_{\mathcal{N}_i,\tau}{\cup}{s_{i,\tau}},a_{\mathcal{N}_i,\tau-1})$. $\beta$ is the coefficient used to calculate entropy loss.

Similarly, to update the critic, the loss function is:
\begin{equation}
\label{valueloss}
\mathcal{L}(w_i)=\frac{1}{|B|}\sum_{\tau\in{B}}(\hat{R}_{i,\tau}-V_{w_i}(s_{i,\tau},a_{\mathcal{N}_i,\tau}))^2.
\end{equation}
Here, $\{(s_{i,\tau}, a_{i,\tau}, s_{i,\tau+1}, r_{i,\tau})\}_{i\in{\mathcal{V}},\tau{\in}B}$ constructs the reply buffer containing experience trajectory. The sampled action value is 
\begin{equation}
\begin{split}
\hat{R}_{i,\tau}=&\sum_{\tau'=\tau}^{|{B}|-1}{\gamma}^{\tau'-\tau}(\sum_{j\in{\mathcal{V}}}\alpha^{d_{i,j}}{r_{j,\tau'}})+\\
&\gamma^{|{B}|-\tau}V_{w_i^-}(s_{\mathcal{N}_i,\tau}{\cup}{s_{i,\tau}},a_{\mathcal{N}_i,\tau-1}).
\end{split}
\end{equation}

\begin{figure*}[!t]
\centering
\includegraphics[width=5in]{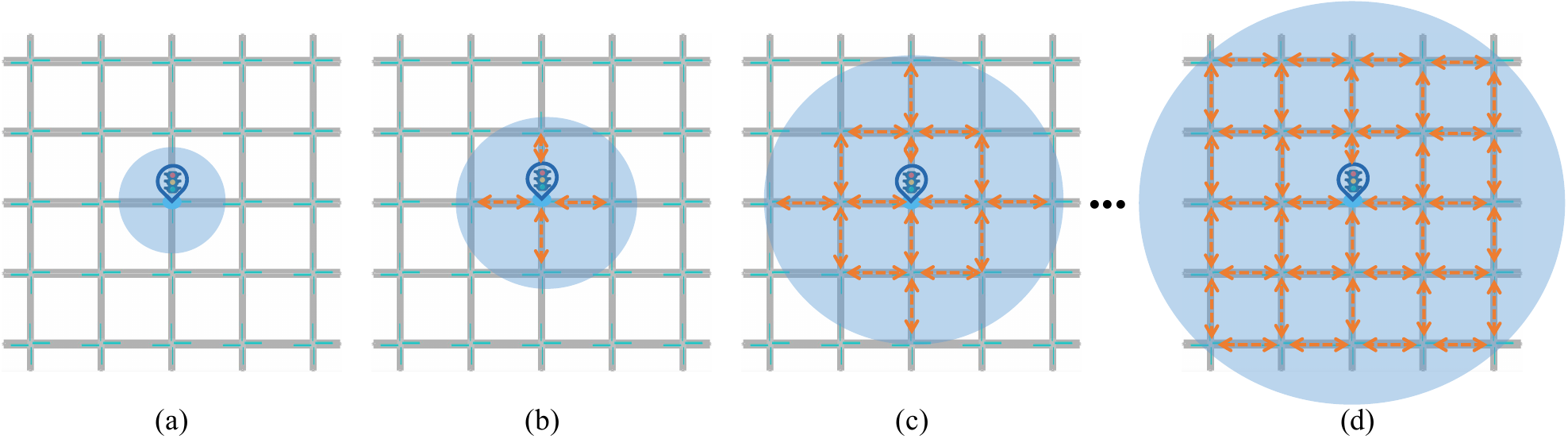}
\caption{{Illustration of information aggregation by selecting an agent $v_i\in\mathcal{V}$ and showing the successive local exchanges with its neighbors. 
(a): Record local observations and build $\mathbf y^{Agg}_{i,t}=[\mathbf y^0_{t}]_i$.
(b): For one-hop neighborhood, record $[\mathbf y^1_{t}]_i$ and build $\mathbf y^{Agg}_{i,t}=[[\mathbf y^0_{t}]_i, [\mathbf y^1_{t}]_i]$. (c): For two-hop neighborhood, record $[\mathbf y^2_{t}]_i$ and build $\mathbf y^{Agg}_{i,t}=[[\mathbf y^0_{t}]_i, [\mathbf y^1_{t}]_i, [\mathbf y^2_{t}]_i$]. (d): For $K-1$-hop neighborhood, record $[\mathbf y^{K-1}_{t}]_i$ and build $\mathbf y^{Agg}_{i,t}=[[\mathbf y^0_{t}]_i, [\mathbf y^1_{t}]_i, [\mathbf y^2_{t}]_i,...,[\mathbf y^{K-1}_{t}]_i]$. Finally, $\mathbf y^{Agg}_{i,t}$ will exhibit a regular structure.}}
\label{AggregationParadigm}
\end{figure*}
 
\subsection{Algorithm Design}

We design a new MARL algorithm for the system, named Aggregated A2C (Agg-A2C). Agg-A2C resorts to the excellent performance of graph learning in capturing 
correlation-related knowledge from unstructured data. {It also improves the scalability of the traditional MARL framework by incorporating graph-based information aggregation into the reinforcement learning pipeline. To improve the training performance of Agg-A2C, we replace the matrix multiplications of GRU with graph convolution and formulate GRU-based graph convolution operation.}

\subsubsection{GRU}
Recurrent Neural Networks (RNNs) represent a kind of neural network for machine learning tasks by handling variable-length sequences. The recurrent hidden state plays a crucial role in the prediction tasks of RNNs by retaining historical information of the neural network. The calculation of recurrent hidden state at time $t$, i.e., $\mathbf h(t)$, is based on $\mathbf h(t-1)$ and current input.

Specifically, considering an input sequence $\boldsymbol x=(\boldsymbol x_1, \boldsymbol x_2,...,\boldsymbol x_T)$, an RNN is used to output $\boldsymbol y=(y_1, y_2,...,y_T)$. The output of the RNN for the next element of the sequence is a conditional probability distribution, i.e., 
\begin{equation}
p(\boldsymbol x_t|\boldsymbol x_1,\boldsymbol x_1,...,\boldsymbol x_{t-1})=p(h_t).
\end{equation}
{The traditional way to calculate recurrent hidden unit at time $t$ is: 
\begin{equation}
\mathbf h_t=g(\mathcal W{\boldsymbol x_t}+U{\mathbf h_{t-1}}).
\end{equation}
At the initial time instant $t=0$, the recurrent hidden unit is updated as $\mathbf h_t=0$.} 
 
GRU is a variant of RNNs with 
sophisticated recurrent units \cite{cho2014properties}. {In GRU, the recurrent hidden unit in traditional RNNs is replaced by update gate $\mathfrak{Z}_t^j$. $\mathfrak{Z}_t^j$ is calculated by the activation $h_t^j$ based on the previous activation $h_{t-1}^j$ and candidate activation $\widetilde{h}_{t}^j$. In addition, GRU has a reset gate $\mathfrak{R}_t^j$ to forget the state computed previously.} For a specific GRU unit, the equation set for the definition above is presented as \cite{cho2014properties}:
{
\begin{equation}
\label{gruprocedures}
\begin{aligned}
h_t &= (1-\mathfrak{Z}_t)h_{t-1}+\mathfrak{Z}_t{\widetilde{h}_{t}}\\
\mathfrak{Z}_t &= \sigma(\mathcal W_\mathfrak{Z}\boldsymbol{x}_t+U_\mathfrak{Z}\mathbf{h}_{t-1})\\
\widetilde{h}_{t}&={\rm tanh}(\mathcal W\boldsymbol{x}_t+U(\mathbf{\mathfrak{R}}_t{\odot}\mathbf{h}_{t-1})) \\
\mathfrak{R}_t &= \sigma(\mathcal W_\mathfrak{R}\boldsymbol{x}_t+U_\mathfrak{R}\mathbf{h}_{t-1})
\end{aligned}
\end{equation}}



   
\subsubsection{Agg-A2C}
{Before depicting the newly proposed Agg-A2C algorithm, we first illustrate a new graph convolution module.}

Recall that a basic graph convolution structure is formed in \ref{graphconvolutionstructure} as a template, we then depict its instantiation under the context of GRU. {That is, the new graph convolution module incorporates the graph convolution in (\ref{graphconvolutionmultiple}) into the traditional GRU unit in (\ref{gruprocedures}) by replacing the matrix multiplications with the diffusion convolution.
As such, the new GRU unit becomes 
\begin{equation}
\label{modifiedgru}
\begin{aligned}
h_t &= (1-\mathfrak{Z}_t)h_{t-1}+\mathfrak{Z}_t{\widetilde{h}_{t}},\\
\mathfrak{Z}_t &=   \sigma(\boldsymbol{\Theta}_\mathfrak{Z}[\boldsymbol{x}_t,\mathbf{h}_{t-1}]+\mathbf{b}_\mathfrak{Z}),\\
\widetilde{h}_{t}&={\rm tanh}(\boldsymbol{\Theta}_{\widetilde{h}}[\boldsymbol{x}_t,(\mathbf{\mathfrak{Z}}_t{\odot}\mathbf{h}_{t-1})]+\mathbf{b}_{\widetilde{h}}), \\
\mathfrak{R}_t &= \sigma(\boldsymbol{\Theta}_\mathfrak{R}[\boldsymbol{x}_t,\mathbf{h}_{t-1}]+\mathbf{b}_{\mathfrak{Z}}).
\end{aligned}
\end{equation}}
Compared with original graph convolution operation defined in (\ref{fixedaggregation}-\ref{graphconvolutionmultiple}), the modified GRU unit in (\ref{modifiedgru}) extends the filter taps $\Psi_l^{fg}$ into 
the combination of weight and bias, i.e., {adding three filters with parameter sets $\boldsymbol{\Theta}_\mathfrak{Z}$, $\boldsymbol{\Theta}_{\widetilde{h}}$, and $\boldsymbol{\Theta}_\mathfrak{R}$, and three biases, i.e., $\mathbf{b}_\mathfrak{Z}$, $\mathbf{b}_{\widetilde{h}}$, and $\mathbf{b}_{\mathfrak{R}}$. }

Based on information aggregation and the graph convolution module, the procedures of Agg-A2C under the MARL framework are illustrated in Algorithm $1$. $T$ is the episode horizon in the training process. $|B|$ denotes the batch size in the replay buffer. $\eta_{w}$ and $\eta_{\theta}$ are the learning rates of critic and actor, respectively. The training process of Agg-A2C is decomposed into six procedures. 
{Firstly, information aggregation is executed so that graph signals can be constructed locally through immediate communications (lines $3-7$).} Each agent then explores experience in lines $8-12$ and samples actions after updating local policy. {Each agent obtains a local reward and observes a new state after criticizing the action by calculating value and executing an action in lines $13-17$. After that, the batch in the replay buffer is updated (line $18$). The explore process will not be terminated until enough experiences are collected or a stop condition is reached.} Initialization is carried out at the end of each episode (lines $20-22$). Both actor and critic are updated in lines $23-31$ where the gradient-based optimizer is adopted.

{Particularly, Agg-A2C preprocess the information at each agent before information aggregation by adding an encoder (line $5$). }
The encoder includes two separate fully-connected layers with $64$ units to encode observation information. {The outputs are first concatenated by a concat function and then input into the graph convolution module.} In the graph convolution module, two GRU layers are connected by a ReLU function. {During the graph convolution operation, the number of time slots aligns with the length of the traffic aggregation sequence constructed locally.}
The final output layer outputs actions and values by setting the activation function of actor and critic as softmax and linear, respectively. 

{
For brief illustration, Fig. \ref{BlockDiagram} depicts the block diagram of Agg-A2C. In Fig.  \ref{BlockDiagram}, we highlight the connections of three stages within the algorithm from the perspective of agent $i$. Both stage I and stage II constitute the main components of SA-GNN, which can be understood as an embedding process. Two GRU layers in stage II include the new GRU units, which are the instantiation of the graph convolution structure of SA-GNN. Stage III employs two distinct activation functions for the actor and critic.
Fig. \ref{NetworkArchitecture} shows the network architecture of Agg-A2C. The relationships between agents are implicitly depicted in the recursive information aggregation process. We highlight those relationships in Fig. \ref{AggregationParadigm} using four sub-figures by depicting the step-by-step information aggregation from the perspective of a single agent.}
Note that Algorithm $1$ illustrates a decentralized and synchronous process by assuming synchronous clocks for agents. Agg-A2C also allows asynchronous execution by setting asynchronous operations for agents. For example, a subset of agents is selected to collect information and perform control at each time $t$. It should be satisfied that information exchange is completed within a constant time interval. In addition, each agent maintains periodic model updating and control. In the next section, we present the experimental results under synchronous settings.
  
 
\begin{algorithm}[t]
\label{AggA2C}
\SetAlgoLined
\KwIn{$\alpha$, $\beta$, $\gamma$, $T$, $\vert{B}\vert$, $\eta_w$, $\eta_{\theta}$}
\KwOut{ Decentralized policies $\pi_i$ for i =1,...N}
 \textbf{initialize}: $s_0$, $\pi_{-1}$, $t \leftarrow 0$, $l \leftarrow 1$, $B=\emptyset$ \\
 \Repeat{Stop condition reached}
{
    \For{$i \in \mathcal{V}$}
    {
    \textbf{observe} agent information and construct $\mathbf {O}_{i}(t)$;\\
    \textbf{encode} local information; \\
    \textbf{collect} information from neighbors and construct aggregation signal (\ref{step4_informationagg});
    }
    \For{$i \in \mathcal{V}$}{
        \textbf{execute} graph convolution operation; \\
        \textbf{update} policy $\pi_{i,t}{\leftarrow}\pi_{\theta_i}$;\\
        \textbf{sample} $a_{i,t}$ from $\pi_{i,t}$; \\
    }
    \For{$i \in \mathcal{V}$}
    {
    \textbf{update} value $v_{i,t}{\leftarrow}V_{w_i}$; \\
    \textbf{execute} $a_{i,t}$; \\
    \textbf{receive} local reward $r_{i,t}$ and transfer to new state $s_{i,t+1}$;
        }
    \textbf{update} batch $B$; \\
    $t{\leftarrow}t+1$, $l{\leftarrow}l+1$; \\
    \If {$t=T$}
    {\textbf{initialize} state and policy, $t{\leftarrow}0$;}
    \If {$l=|B|$}
    {
    \For{$i \in \mathcal{V}$}
        {
        \textbf{estimate} $\hat{R}_{i,\tau}, \tau{\in}B$, \\
        \textbf{estimate} $\hat{A}_{i,\tau}, \tau{\in}B$, \\
        \textbf{update} $w_i$ with $\eta_{w}{\Delta}{\mathcal{L}(w_i)}$ \\
        \textbf{update} $\theta_i$ with $\eta_{\theta}{\Delta}{\mathcal{L}(\theta_i)}$
        }
    $B{\leftarrow}\emptyset$, $l{\leftarrow}0$
    }
}
 \caption{Agg-A2C}
\end{algorithm}

\section{Performance Evaluation}
{We conduct extensive experiments to evaluate the performance of the newly proposed Agg-A2C algorithm with state-of-the-art MARL algorithms and a traditional algorithm.} {The experimental results convey two insights into traffic light control problems under the decentralized control framework. The first insight is that graph signals constructed over system topology improve the cooperation of controllers in decentralized control systems by resorting to the excellent learning capacity of graph learning in unstructured data.} This is different from traditional solutions that extend the information range of controllers directly. Agg-A2C provides a cost-effective construction strategy of graph signals by incorporating the diffusion process into information aggregation. The second insight is that the performance gain due to graph signals is higher in large-scale systems, which confirms our original idea that Agg-A2C is robust to system scales.

\subsection{Experimental Settings}
\subsubsection{Traffic System} 
In terms of experimental stringency, we develop two experimental environments with different scales based on traffic simulation data and a real-world traffic dataset, respectively. 

{The experimental environments are implemented in SUMO (Simulation of Urban MObility) by importing synthetic and real-world traffic data.} As an open-sourced, highly portable traffic simulator \cite{krajzewicz2012recent}, SUMO has become popular in the performance evaluation of microscopic traffic.
SUMO integrates the interfaces that detect road conditions (e.g., traffic volume) and sets the phases of traffic lights. After setting the initial vehicle status (e.g., speed, acceleration, driving trips), {SUMO begins simulating the networked control system by controlling traffic lights based on instructions from external algorithms.} 

We first implement an experimental environment in SUMO by setting a total of $25$ intersections, where traffic lights are connected via lanes in a grid way. {Before the simulation, we generate numerous time-variant traffic flows that define the vehicle trips at different times. We set the traffic flows by including various traffic demands, corresponding to the time-varying traffic scenarios in practical systems. Specifically, six traffic flows are
included, i.e., three major flows and three minor flows, where the route of each vehicle is randomly generated during run-time. The major flow holds a dynamic vehicle genera rate among $[0-1100]$ veh./hour. The minor flow generates vehicles at a dynamic rate of $[0-700]$ veh/hour. Each traffic flow is set with a given Origin-Destination (O-D) pair. After $15$ minutes, the volumes of all traffic flows start to decrease, while their opposite flows (with swapped O-D pairs) start to be generated.}
Based on the lanes attached at each traffic light, {we set five available actions for the adjustment of signal phases and each of them is a combination of straight, left-turn, and right-turn phases.} At each time $t$, the state of an individual traffic light can be detected locally by induction-loop detectors. 
 
We then develop an experimental environment in SUMO based on a real-world traffic dataset \cite{codeca2018monaco}. \cite{codeca2018monaco} provides a data statistic report for Monaco City by including the behaviors of various kinds of vehicles, road users, and public transport in practical road networks. {The traffic data in \cite{codeca2018monaco} covers a maximum area of $70$ $km^2$ around the city. We select an area consisting of $30$ intersections with various connected lanes.} {The settings of traffic flows in the experiments with real-world data are different. Specifically, four
traffic flow groups are generated as a multiple of $unit$ flows of $325$ veh./hour, with randomly sampled O-D pairs inside given areas.}

{
In our experiments, each traffic light is controlled by the actions decided by the corresponding agent. To ensure the smooth operation of the traffic systems, regular interactions between agents and traffic lights are required. In our experiments, we set $\Delta{t}$ as a fixed interaction interval of $5$ seconds. 
That is, every $5$ seconds, each agent outputs actions and the corresponding traffic signal accordingly adjusts its signal phase. After adjusting all traffic lights, the system is simulated for a duration of $5$ seconds until the next decision-making phase. In each episode, the total interaction number is set as $720$, resulting in a time horizon of $3600$ seconds. 
In addition, we also set a time slot of $2$ seconds as the guard interval after each signal phase switch to ensure traffic safety, corresponding to real-world traffic systems.
}


\subsubsection{Baseline Algorithms}

We select multiple start-of-the-art algorithms and apply them in traffic light control to make a performance comparison. 
\begin{itemize}
\item\textbf{IA2C:}{ IA2C is an MARL algorithm developed in \cite{chu2019multi} through replacing the Q-learning module of IQL \cite{tan1993multi}\cite{tesauro2003extending} by a A2C module. It is also a decentralized control method since each agent is allowed to learn its own policy and also update value function locally. The learning process of IA2C is based on an assumption that global reward and state could be observed at each agent.} 
\item\textbf{{IA2C with FingerPrint (IA2C{\_}FP):}}{ IA2C{\_}FP is proposed in \cite{foerster2017stabilising} to combine fingerprints of neighbors into local value function. Specifically, each agent learns local policy conditioned on the policy estimation by observing the behaviors of other agents. {IA2C{\_}FP has the potential to avoid the nonstationarity problem if each agent only knows the local state, and nonstationary transition problem in experience replay.}
}
\item\textbf{{MA2C with NeurComm (MA2C{\_}NC):}}{ MA2C{\_}NC is also a MARL algorithm with a core module of differentiable information exchange strategy (NeurComm), which improves the information acquisition of each agent via information sharing (such as state and behaviors). 
The shared information is spatially discounted and thus MA2C{\_}NC is able to balance information acquisition and information redundancy at each agent. An LSTM-based model is used to process the combined information and then output control actions.
}
\item\textbf{{MA2C with Consensus Networks (MA2C{\_}CNet)}:} MA2C{\_}CNet\cite{zhang2018fully} maintains a consensus update step that updates the critic at each agent based on a weighted combination of parameter estimates from the closed neighborhood. 
\item\textbf{{Max-Pressure}:} {Max-Pressure \cite{varaiya2013max}\cite{varaiya2013maxpartc} is a popular heuristic algorithm in traffic signal control. With Max-Pressure, each agent selects the action with the maximum pressure of each \emph{traffic movement}. The traffic movement (see Definition $3.2$ in \cite{wei2019presslight}) indicates the traffic status in an intersection by collecting vehicles across the intersection from incoming lanes to outgoing lanes. Thus, the key principle of Max-Pressure is that it greedily takes actions to maximize the throughput of road networks, under the assumption that the downstream lanes have unlimited capacity. Notably, the results of Max-Pressure exhibit only in the part of performance evaluation.
} 
\item\textbf{Agg-A2C:} {We implement the newly proposed algorithm illustrated in Algorithm $1$ to make performance comparison.}
\end{itemize}

\begin{figure*}[htbp]
\centering
\hspace{-1.9cm}
\subfigure[On simulation data]{
\begin{minipage}[t]{0.5\linewidth}
\label{trainingonsumo_train}
\centering
\includegraphics[width=3in]{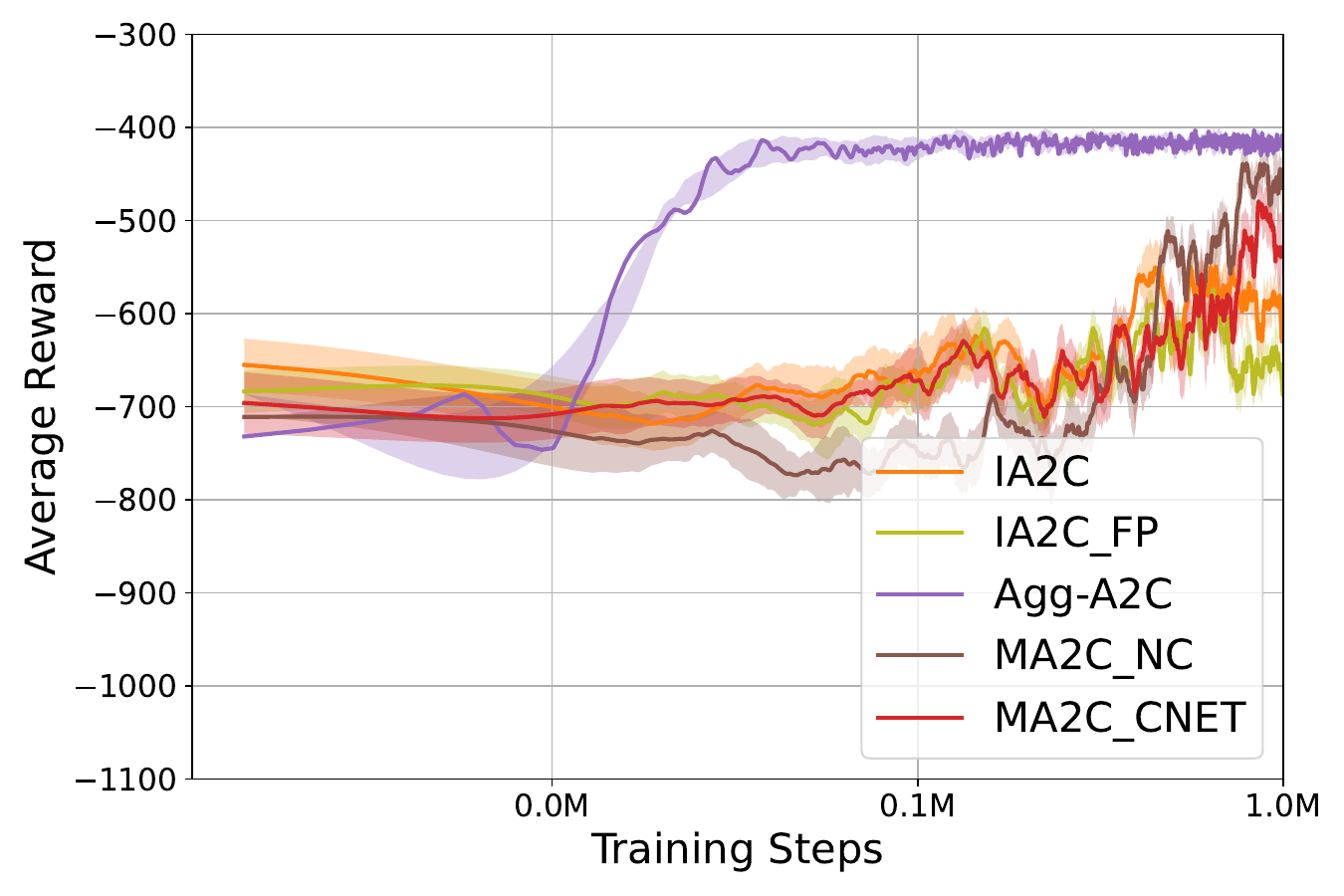}
\end{minipage}%
}%
\hspace{-0.3cm}
\subfigure[On real-world data]{
\begin{minipage}[t]{0.40\linewidth}
\label{trainingonrealworld_train}
\centering
\includegraphics[width=3in]{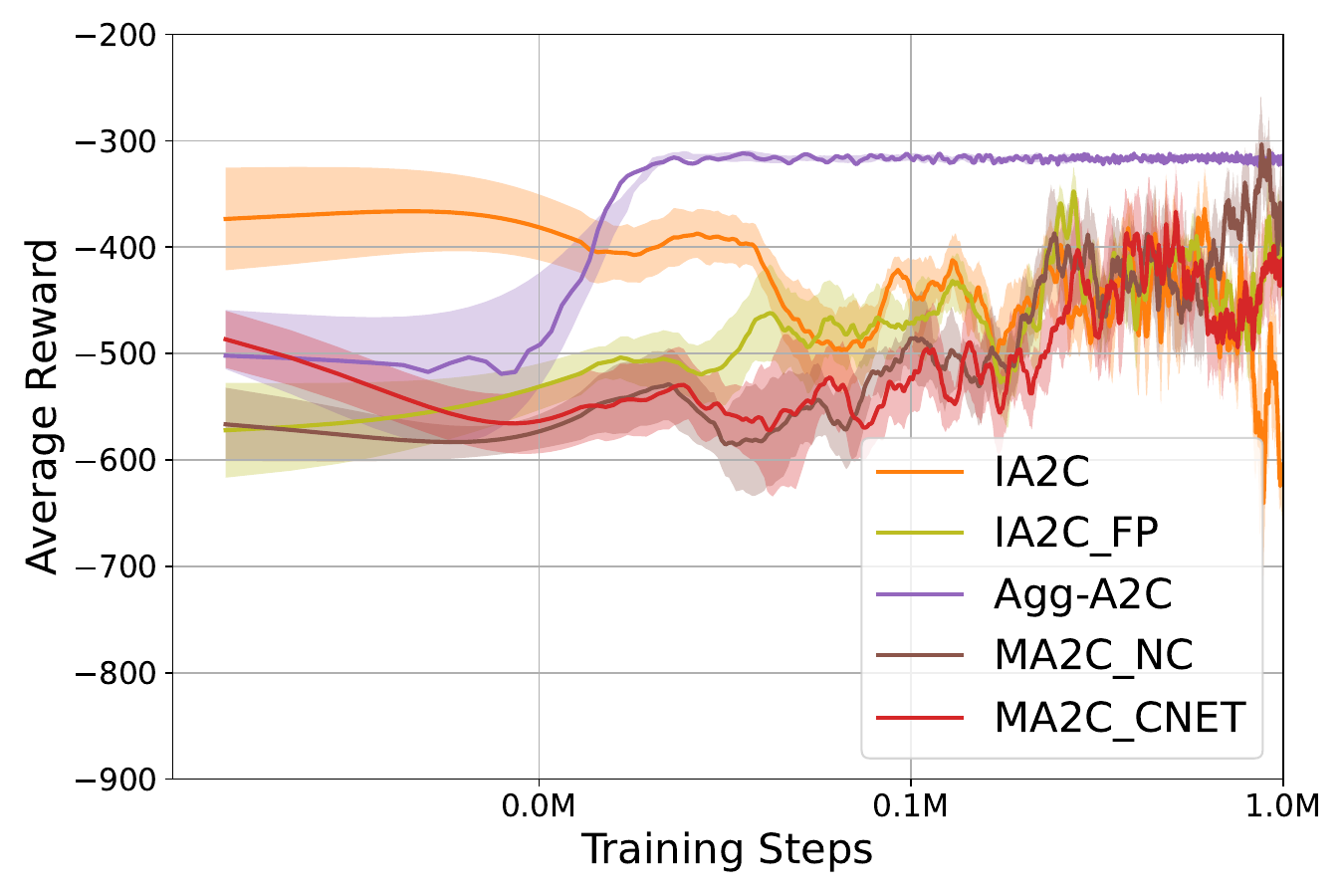}
\end{minipage}%
}%
\centering
\vspace{-0.2cm}
\caption{Training performance in different experimental environments.}
\label{Trainingresults}
\vspace{-0.4cm}
\end{figure*}
\begin{figure*}[htbp]
\centering
\hspace{-1.8cm}
\subfigure[On simulation data]{
\begin{minipage}[t]{0.5\linewidth}
\label{evaluationonsumo_queue}
\centering
\includegraphics[width=2.7in]{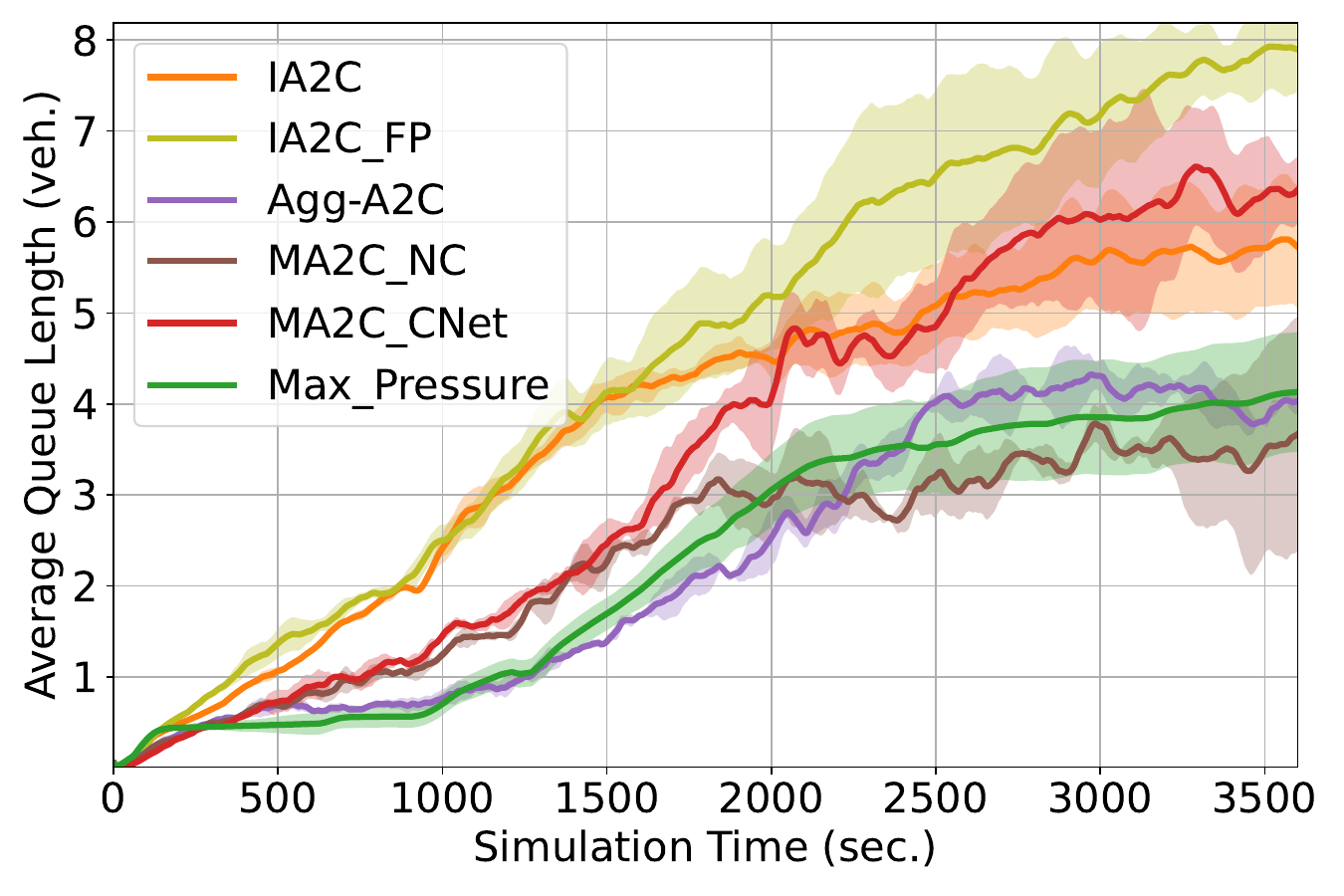}
\end{minipage}%
}%
\subfigure[On real-world data]{
\begin{minipage}[t]{0.40\linewidth}
\label{evaluationonrealworld_queue}
\centering
\includegraphics[width=2.7in]{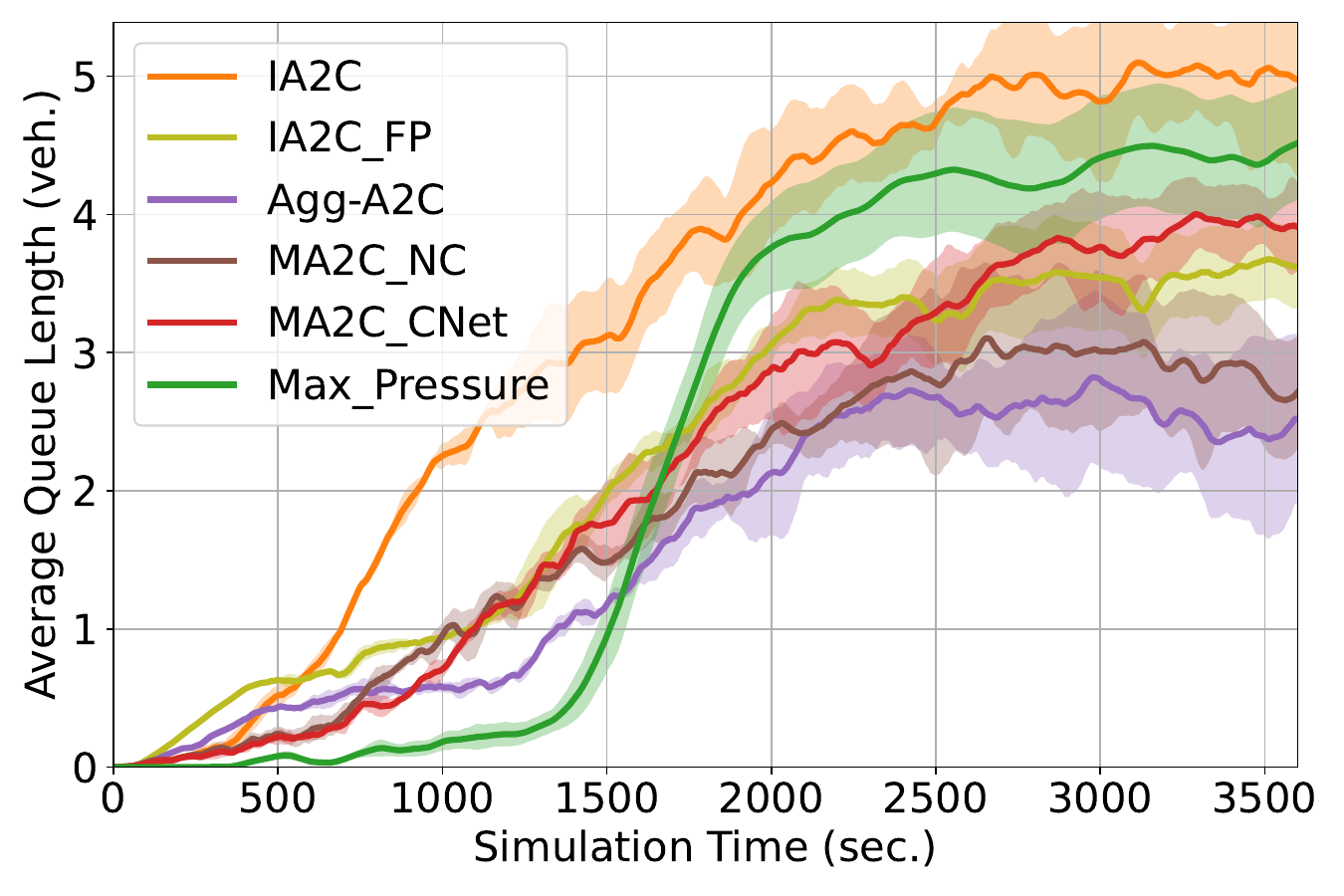}
\end{minipage}%
}%
\centering
\vspace{-0.2cm}
\caption{Queue performance in different experimental environments.}
\label{evaluationqueue}
\vspace{-0.4cm}
\end{figure*}


\subsubsection{Training \& Evaluation} 

Unless indicated otherwise, the details in both training and evaluation phrases are illustrated below.

We train all the algorithms in $1$M steps, corresponding to $1388$ hours of experiments in a real-world traffic system. Each episode lasts for $1$ hour. In each episode, the experimental environment is initialized using a different random seed to enable the diversity of observations, while different algorithms are trained with the same seed. 
We set the initial learning rate of the actor as $5\times10^{-4}$ and the critic as $2.5\times10^{-4}$. Both of them are adaptively adjusted during training. The popular optimizer Adam is adopted. The loss function for the actor and the critic is defined in (\ref{policyloss}) and (\ref{valueloss}), respectively. The batch size in the replay buffer is set as $|B|=120$. During training, three parameters in the algorithm, i.e., $\alpha$, $\beta$, and $\gamma$, are set as $0.75$, $0.01$ and $0.99$, respectively. We implement the experimental environments and all algorithms in TensorFlow. All results will be presented as average values and standard deviation calculated over episodes, denoted by solid lines and shadow areas, respectively.

In the evaluation phase, we test all the trained algorithms through total $20$ episodes and present the mean results in a single episode. Different random seeds are set to initialize the experimental environments for each episode, which is same with the training phase. To fully evaluate the impact of different algorithms on driving experiences and efficiency, we define four evaluation metrics and present their formal definitions in the following subsection. 

{
\subsubsection{Performance Metrics}

We conduct a performance comparison between Agg-A2C and various baselines by illustrating the results obtained from both the training and evaluation phases, outlined in \ref{trainingresults} and \ref{evaluationresults}, respectively. In addition to recording the results related to rewards, we adopt four metrics to evaluate the performance comparison, i.e., average queue length, average car speed, average trip delay, and average intersection delay, which are specifically defined as follows.

\begin{itemize}
\item\textbf{Average Queue Length:}{
As a practical metric that significantly affects the driving experiences of users, queue length refers to the count of vehicles waiting at each incoming lane of a traffic light. By considering a total $N$ intersections, the average queue length is defined as:
\begin{equation}
\label{queueaverage}
\overline{\rm{queue}}_t = \frac{1}{N}\sum_{i=1}^{N}{\rm{queue}}_{i,t},
\end{equation}
where ${\rm{queue}}_{i,t}$ is the queue length along each incoming lane at time $t$, defined as:
\begin{equation}
\label{queuepreagent}
{\rm{queue}}_{i,t} = \sum_{g\in{\mathcal{G}_{ji}},{ji}\in\mathcal{E}} {\rm{veh.}}_{t+\Delta{t}}[g].
\end{equation}
Equation (\ref{queuepreagent}) refers to the total number of vehicles waiting at all incoming lanes $\mathcal{G}_{ji}$ of intersection $i$. In experiments, the number of vehicles is recorded within $\Delta{t}$, corresponding to the information collection duration in experimental settings.
}
\item\textbf{Average Car Speed}{ refers to the mean velocity of all vehicles in the system, which is also a critical factor that indicates the quality of travel experiences and performance of traffic flow. The specific definition is shown as follows:
\begin{equation}
\overline{\rm{speed}}_t = \frac{1}{N_{veh.}[t]}\sum_{i=1}^{N_{veh.}[t]}{\rm{speed}}_{i,t},
\end{equation}
where ${N_{veh.}[t]}$ is the instantaneous number of vehicles in the system while ${\rm{speed}}_{i,t}$ is the velocity of vehicle $i$ at time $t$.
}

\item\textbf{Average Trip Delay}{ is defined as the mean trip time of all vehicles in the system, also indicating the quality of travel experiences and performance of traffic flow. The formal definition is shown as follows:
\begin{equation}
\overline{\rm{delay}}_{{trip}} = \frac{1}{N_{veh.}}\sum_{i=1}^{N_{veh.}}t_{i,ed}-t_{i,st},
\end{equation}
where $t_{i,ed}$ and $t_{i,st}$ denote the ending time and starting time, respectively, of the trip of vehicle $i$. $N_{veh.}$ is the total number of vehicles in the system.
}
\item\textbf{Average Intersection Delay}{ is a simplified metric of \textbf{Average Trip Delay}, calculated by aggregating the time each vehicle spends traversing intersections on its trip. It is defined as: 
\begin{equation}
\overline{\rm{delay}}_{{int.}} = \frac{1}{N_{veh.}[t]}\sum_{i=1}^{N_{veh.}[t]}\sum_{int.\in{\mathcal N_{veh_i}}}t_{i,int.,ed}-t_{i,int.,st},
\end{equation}
where ${\mathcal N_{veh_i}}$ represents set of 
intersections encountered by vehicle $i$ during its trip. 
}
\end{itemize}
}

\subsection{Results}
    
\subsubsection{Training Results}
\label{trainingresults}
Fig.~\ref{Trainingresults} compares the training performance of different algorithms on simulation data and real-world datasets, respectively. Average episode reward denotes the longer-term time average reward along the time scale of each episode. From two figures in Fig.~\ref{Trainingresults}, it is observed that the proposed Agg-A2C converges faster compared with other MARL algorithms. Specifically, both Agg-A2C and MA2C{\_}NC finally converge to a similar performance level while the other three algorithms obtain a lower convergence level. {This is because both Agg-A2C and MA2C{\_}NC enhance information acquisition of agents and thus can stabilize the training process of MARL with partial observation.}
Compared with MA2C{\_}NC, Agg-A2C achieves a faster convergence rate, which is because it incorporates constructed graph signals into a diffusion convolution module rather than directly combining them using a spatially discounted way. 

\begin{figure*}[!htbp]
\centering
\hspace{-1.8cm}
\subfigure[On simulation data]{
\begin{minipage}[t]{0.5\linewidth}
\label{evaluationonsumo_stepreward}
\centering
\includegraphics[width=3in]{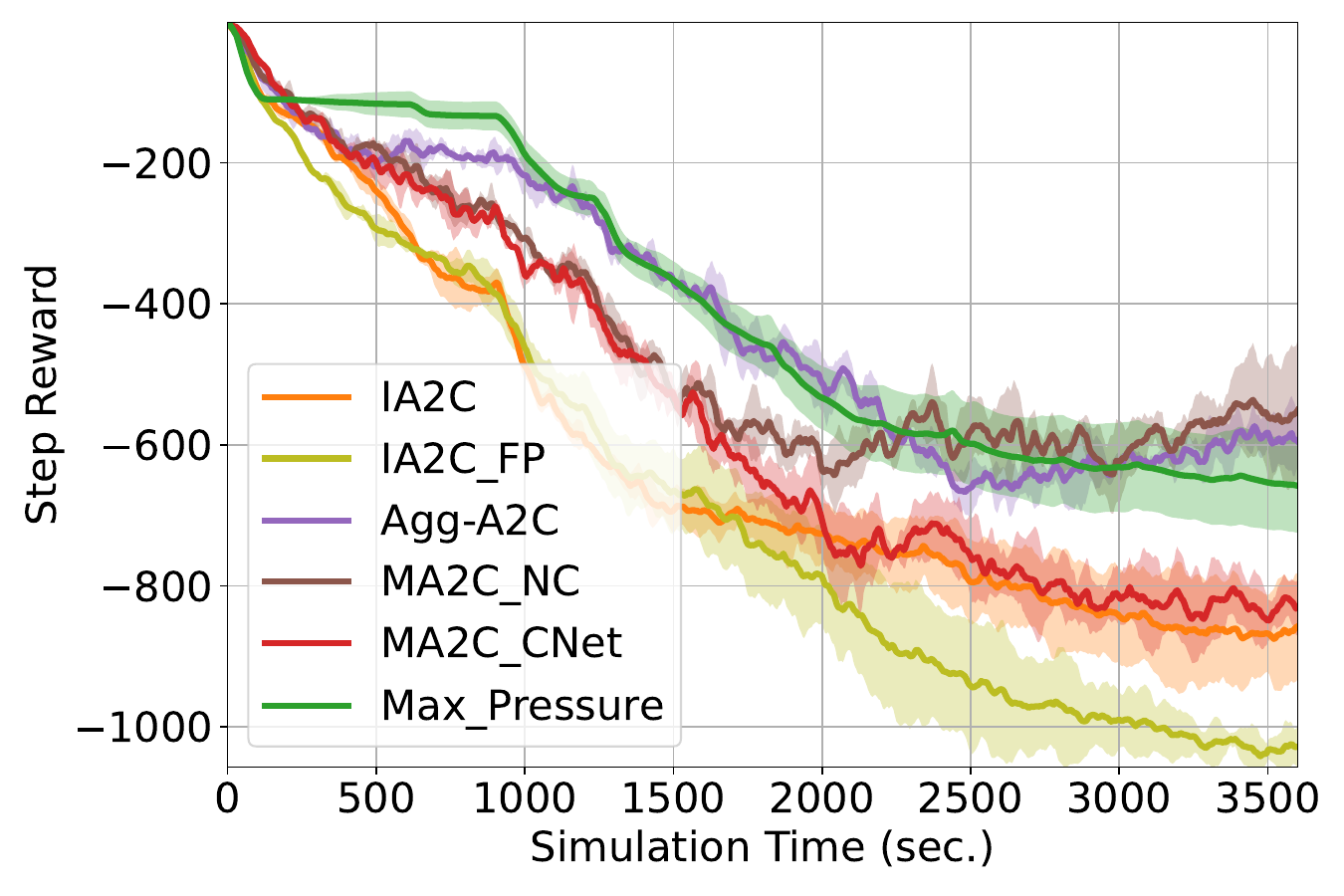}
\end{minipage}%
}%
\subfigure[On real-world data]{
\begin{minipage}[t]{0.45\linewidth}
\label{evaluationonrealworld_stepreward}
\centering
\includegraphics[width=3in]{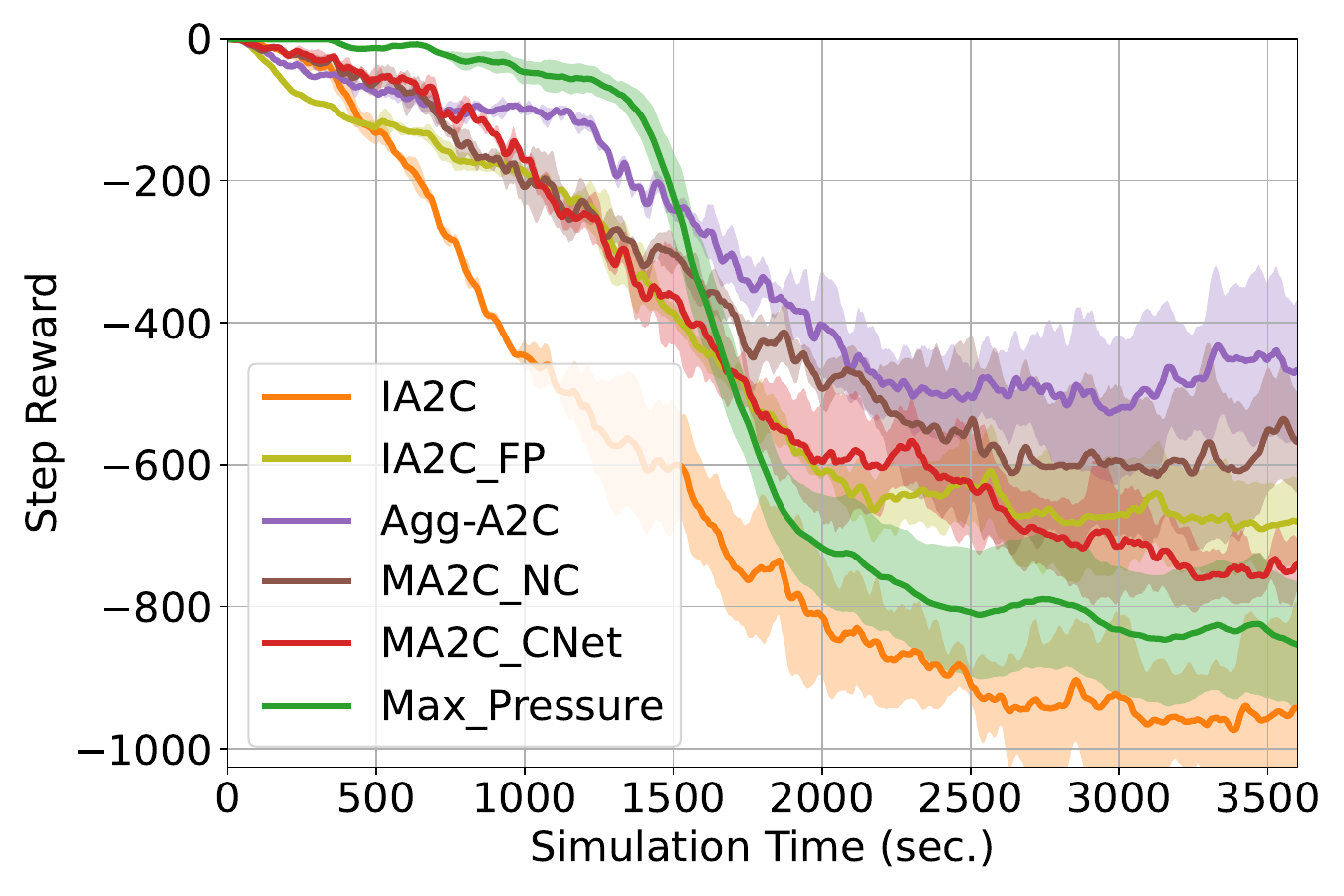}
\end{minipage}%
}%
\centering
\vspace{-0.2cm}
\caption{Reward performance in different experimental environments.}
\label{Evaluationstepreward}
\vspace{-0.6cm}
\end{figure*}
\begin{figure*}[htbp]
\centering
\hspace{-1.6cm}
\subfigure[On simulation data]{
\begin{minipage}[t]{0.5\linewidth}
\label{evaluationonsumo_speed}
\centering
\includegraphics[width=2.7in]{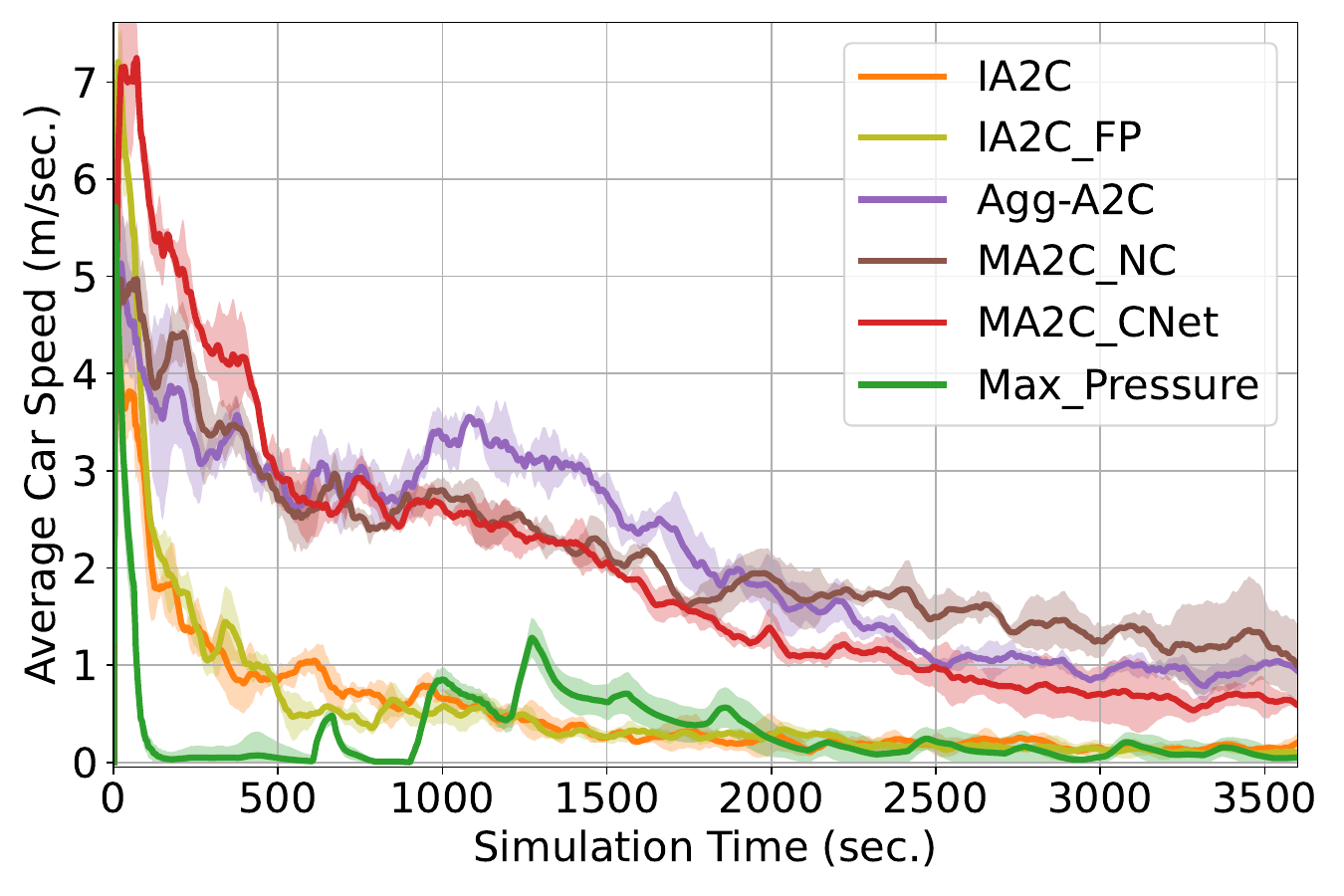}
\end{minipage}%
}%
\subfigure[On real-world data]{
\begin{minipage}[t]{0.40\linewidth}
\label{evaluationonrealworld_speed}
\centering
\includegraphics[width=2.7in]{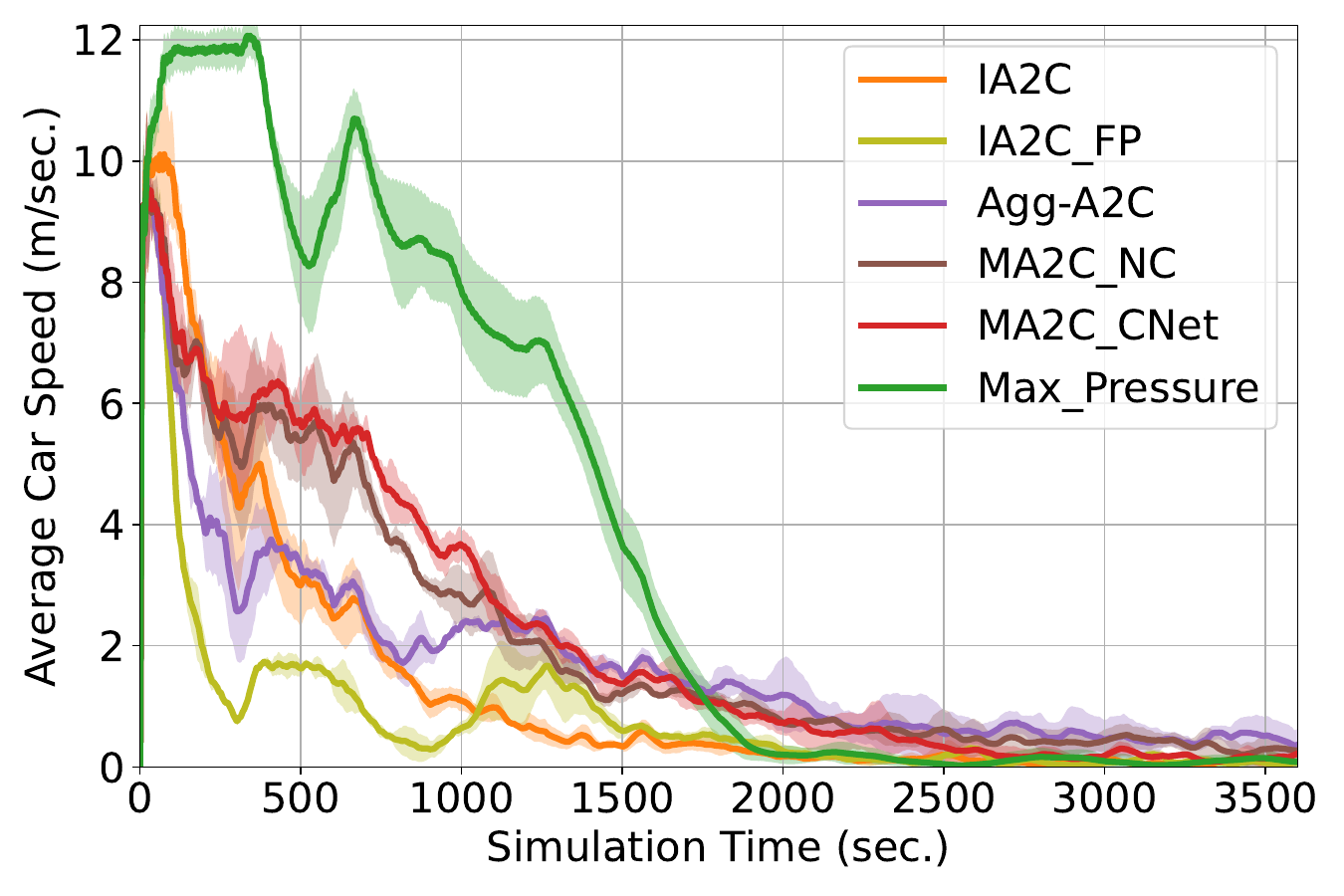}
\end{minipage}%
}%
\centering
\vspace{-0.2cm}
\caption{Speed statistics in different experimental environments.}
\label{Evaluationspeed}
\vspace{-0.6cm}
\end{figure*}
 
\subsubsection{Evaluation Results}
\label{evaluationresults}
\begin{figure*}[htbp]
\centering
\hspace{-1.8cm}
\subfigure[On simulation data]{
\begin{minipage}[t]{0.5\linewidth}
\label{evaluationonsumo_tripwait}
\centering
\includegraphics[width=2.9in]{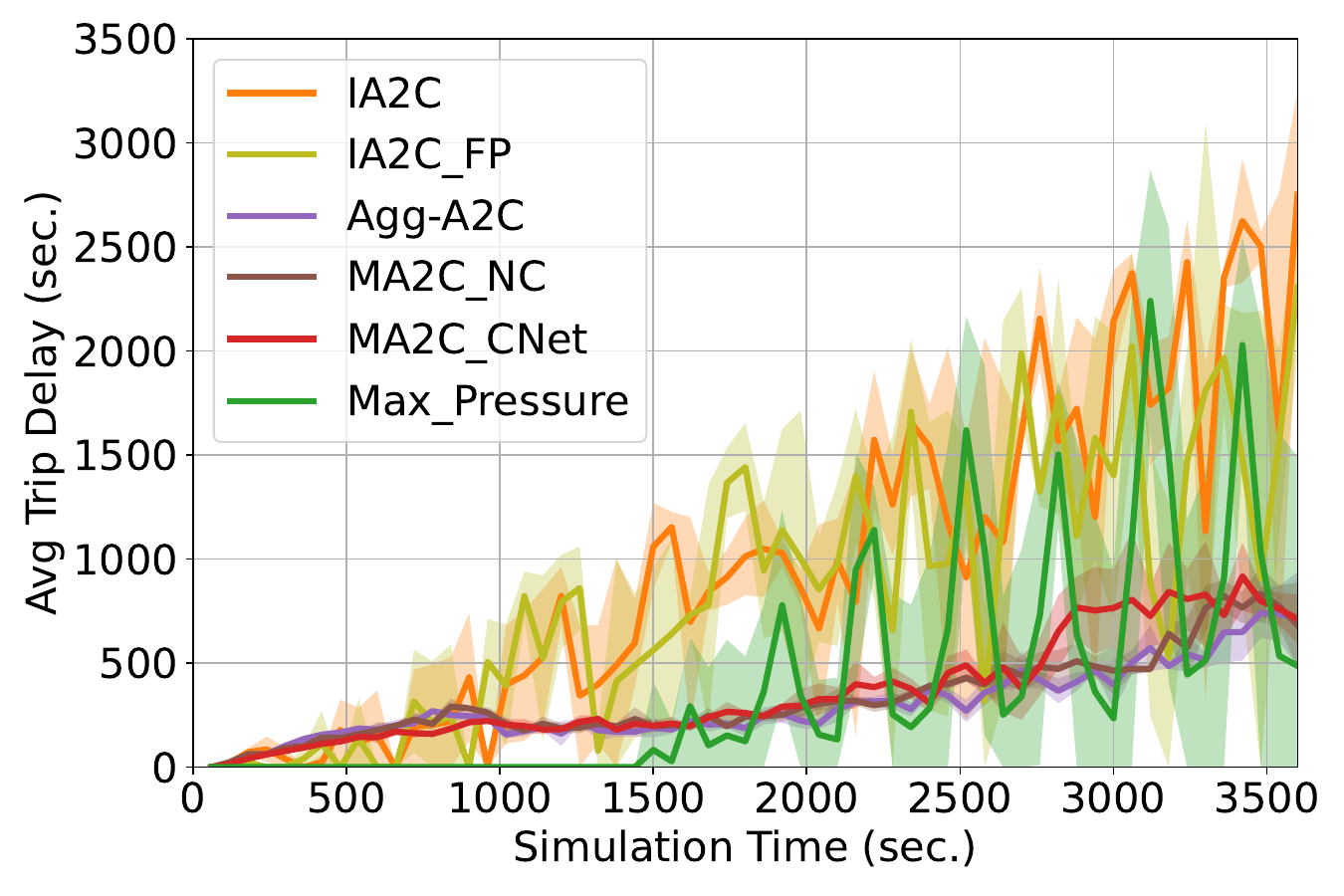}
\end{minipage}%
}%
\subfigure[On real-world data]{
\begin{minipage}[t]{0.40\linewidth}
\label{evaluationonrealworld_tripwait}
\centering
\includegraphics[width=2.9in]{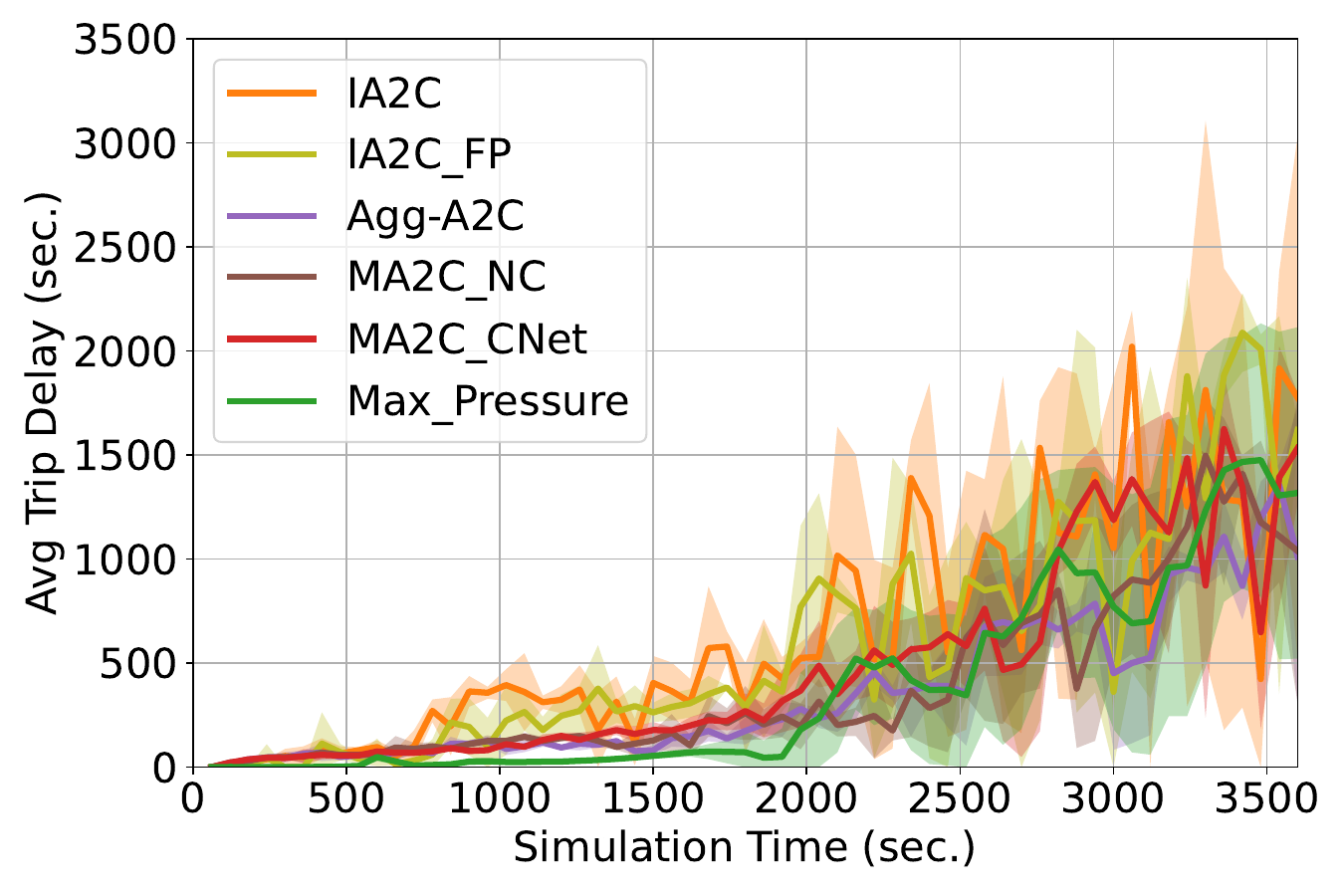}
\end{minipage}%
}%
\centering
\vspace{-0.2cm}
\caption{Trip time in different experimental environments.}
\label{Evaluationtriptime}
\vspace{-0.6cm}
\end{figure*}

\begin{figure*}[htbp]
\centering
\hspace{-1.8cm}
\subfigure[On simulation data]{
\begin{minipage}[t]{0.5\linewidth}
\label{evaluationonsumo_wait}
\centering
\includegraphics[width=2.8in]{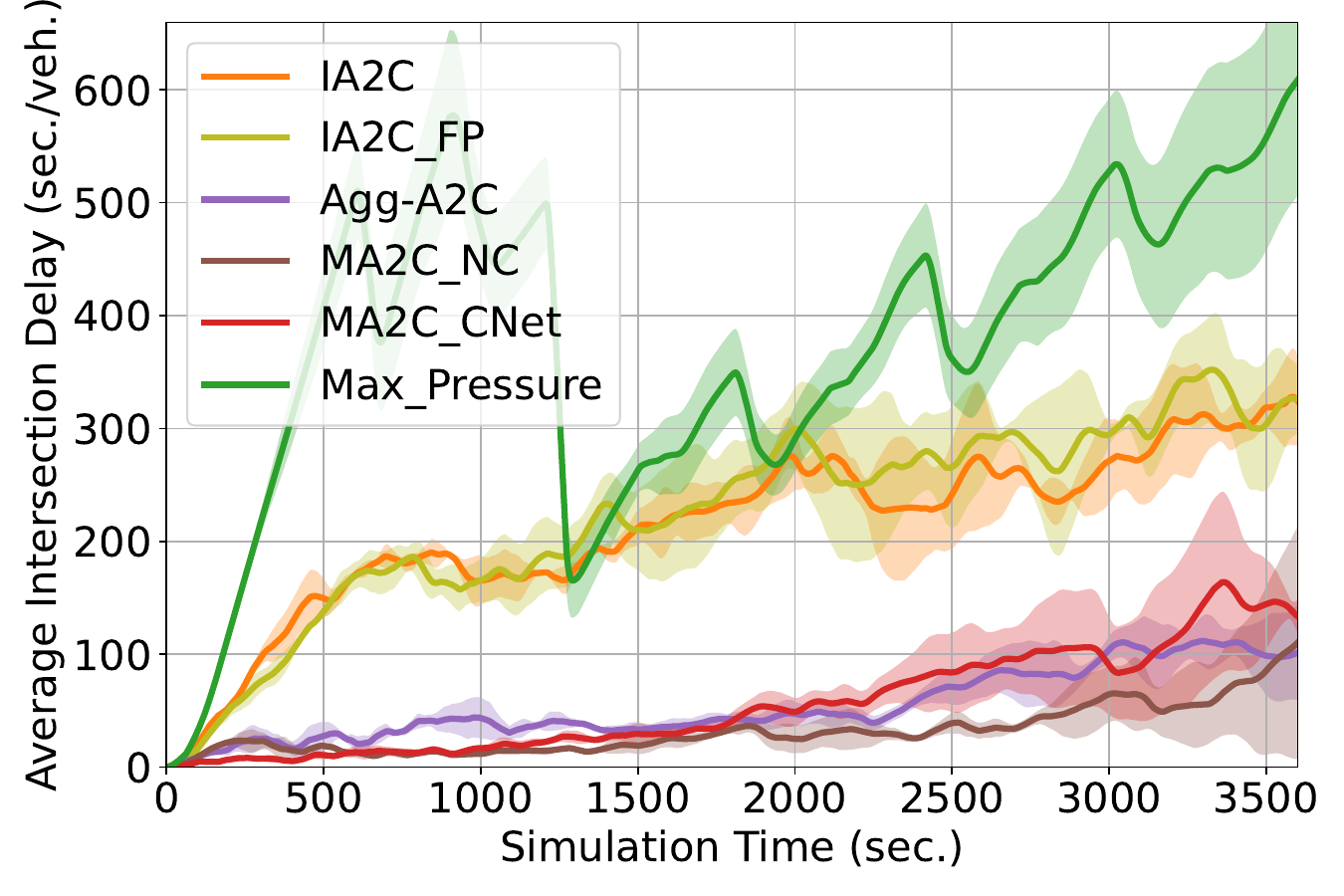}
\end{minipage}%
}%
\subfigure[On real-world data]{
\begin{minipage}[t]{0.40\linewidth}
\label{evaluationonrealworld_wait}
\centering
\includegraphics[width=2.8in]{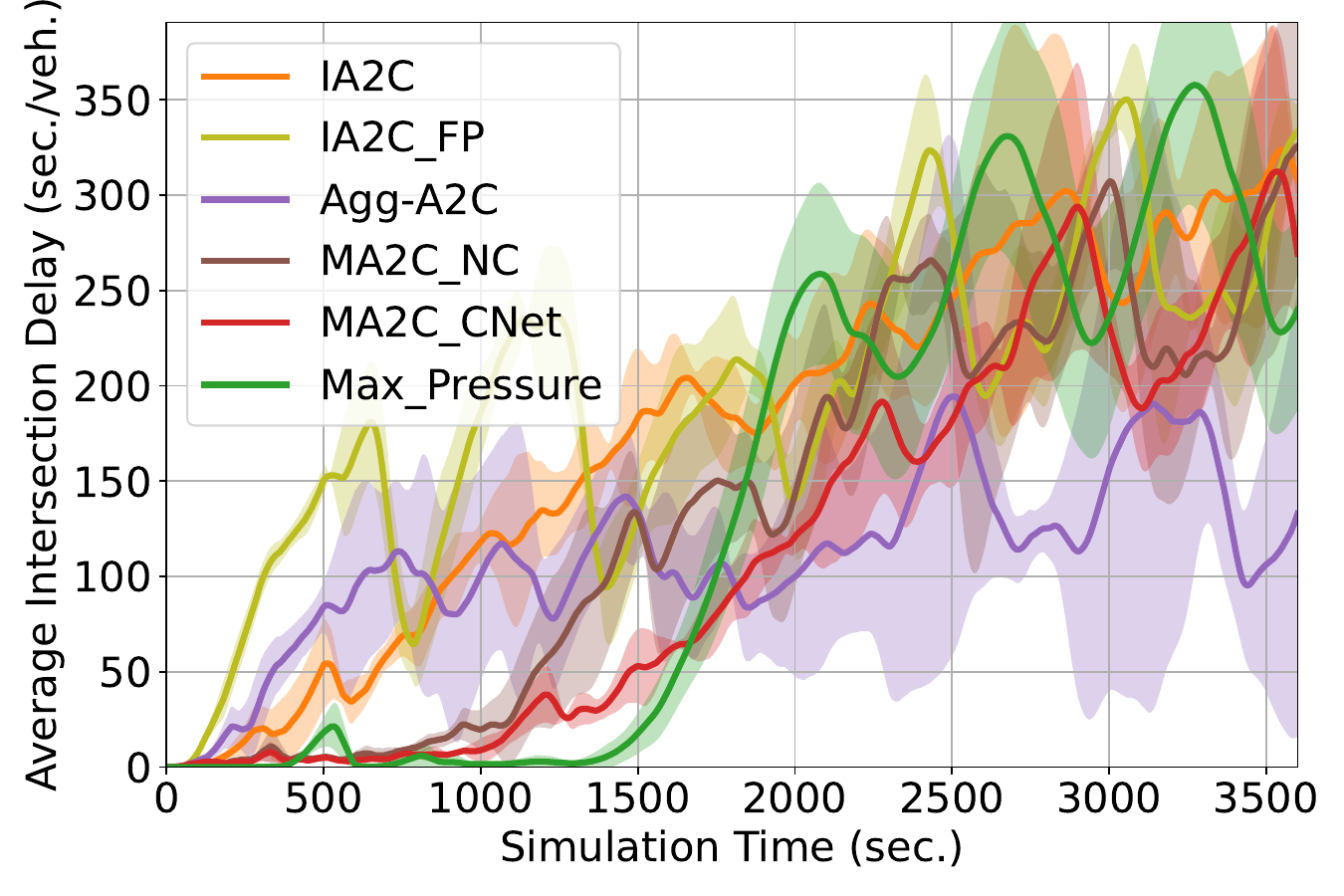}
\end{minipage}%
}%
\centering
\vspace{-0.2cm}
\caption{Waiting results at intersections in different experimental environments.}
\label{Evaluationintersectiondelay}
\vspace{-0.6cm}
\end{figure*}

{Fig.~\ref{evaluationqueue}-Fig.~\ref{Evaluationintersectiondelay} present the evaluation results in two experimental environments with the performance comparison of different algorithms.} 

Specifically, {Fig.~\ref{evaluationqueue} shows the statistics results of average queue length at each simulation step in two experimental environments. A natural phenomenon in practical traffic systems is that the queue length will accumulate gradually if the traffic lights are not controlled properly.} In Fig.~\ref{evaluationqueue}, the lower level of average queue length indicates a better traffic management performance achieved by the corresponding algorithm. From Fig.~\ref{evaluationqueue} we observe that the average queue length tends to be stable finally based on the decisions of those algorithms. {The performance advantage of both Agg-A2C and MA2C{\_}NC in both experimental environments is also observed.} Notably, in terms of the final stable queue state, MA2C{\_}NC outperforms Agg-A2C in the experiments on the simulation dataset while a reverse conclusion is drawn in the experiments on the real-world dataset. {In addition, we can observe that MaxPressure obtains similar performance compared to our proposed Agg-A2C algorithm in the synthetic experiments. However, in the experiments with real-world data, the performance gap between Max-Pressure and Agg-A2C is evident.}
This validates that Agg-A2C is more suitable in practical systems with complex topology since the simulation dataset is acquired from a grid simulation environment. 




{
Fig.~\ref{Evaluationstepreward} shows the results of reward collected at various experimental steps within two experimental environments.} Different from the time-average reward in Fig.~\ref{Trainingresults}, Fig.~\ref{Evaluationstepreward} shows the step reward at each experiment and thus exhibits a decreasing trend. From Fig.~\ref{evaluationonsumo_stepreward}, it is seen that both Agg-A2C and MA2C{\_}NC achieve better performance compared with MA2C{\_}CNet, IA2C and IA2C{\_}FP while the reward obtained by Agg-A2C and MA2C{\_}NC reaches to a close level finally. According to Fig.~\ref{evaluationonrealworld_stepreward},
IA2C exhibits the worst performance while Agg-A2C is the best among the other four algorithms. The performance gap between Agg-A2C and MA2C{\_}NC is apparent, which is similar to the results shown in Fig.~\ref{evaluationonrealworld_queue}. 
{Consistent with the findings depicted in Fig.~\ref{evaluationqueue}, Max-Pressure performs well in synthetic experiments but exhibits weak results in real-world data experiments.}
Overall, the results in Fig.~\ref{evaluationonsumo_stepreward} further validate the effectiveness of Agg-A2C in decentralized control systems with complex topology. 

Fig.~\ref{Evaluationspeed}-Fig.~\ref{Evaluationintersectiondelay} present the travel experiences of vehicles affected by different algorithms in both experimental environments. {Specifically, Fig.~\ref{Evaluationspeed} shows the statistics results of average car speed}. The larger average speed means that traffic flow is well controlled in the experiment with gradually rising traffic congestion. Fig.~\ref{evaluationonsumo_speed} shows that Agg-A2C, MA2C{\_}NC, and MA2C{\_}CNet finally reach a close performance level and they outperform IA2C and IA2C{\_}FP. However, there is a little performance gap among different algorithms exhibited in Fig.~\ref{evaluationonrealworld_speed}. In addition, we see the low average speed, which is because the traffic system reaches a congestion state. Nevertheless, both Agg-A2C and MA2C{\_}NC still contain the recovery capacity in such a saturated condition. {
The reason why traffic saturates to the congestion is that output (trip
completion) flow naturally increases in the experiments. Specifically, the output flow is impacted by vehicle accumulation. An intuitive finding is that when vehicle density is low, output flow increases as accumulation grows; when the network becomes more saturated, further accumulation will decrease the
output, leading to potential congestion.}

{
Fig.~\ref{Evaluationtriptime} and Fig.~\ref{Evaluationintersectiondelay} show the results of two metrics that influence travel time, i.e., average trip delay and average intersection delay. Specifically, similar to the results in Fig.~\ref{Evaluationspeed}, the performance of travel time reflects the capacities of algorithms in managing traffic flow as well as recovery in congested conditions. From Fig.~\ref{Evaluationtriptime} we see that Agg-A2C presents a superior traffic control capacity since it ensures that vehicles travel in the lowest time}, similar to MA2C{\_}NC and MA2C{\_}CNet. In terms of average intersection delay, 
three algorithms, i.e., Agg-A2C, MA2C{\_}NC, and MA2C{\_}CNet, obtain similar performance in the experiments based on simulation dataset as shown in Fig.~\ref{evaluationonsumo_wait}. {In the experiments based on real-world data, Agg-A2C obtains the best performance and presents an apparent advantage as shown in Fig.~\ref{evaluationonrealworld_wait}. This also validates that Agg-A2C has a better recovery capacity compared with other baseline algorithms in saturated conditions of complex control systems.}

%
 
\section{Conclusions}
We conclude the paper by emphasizing that the foundation
of adaptive decentralized control systems lies in the effective cooperation of agents based on the cost-effective acquisition of environmental information. 
In this work, we proposed to improve MARL by forming a new information aggregation strategy based on the principle of graph learning. By diffusion-aware information aggregation, our new MARL algorithm improves the management efficiency of traffic systems and also enhances travel experiences.
To the best of our knowledge, this is the first study
that explores diffusion graph convolution in MARL-controlled adaptive decentralized control systems. This work reveals that graph learning is applicable in practical systems with complex topology.

{
For future work, it is necessary to extend the graph learning based MARL algorithm into generalized traffic systems by including more elements, such as various vehicles, pedestrians, and different road infrastructure.} In addition, it is interesting to reveal the theoretical basis of graph learning in decentralized control systems, e.g., to explain how the system scale and topology complexity impact the cooperation of agents.

 
 
 


%
 
\bibliographystyle{IEEEtran}
\bibliography{ref} 
\begin{IEEEbiographynophoto}{Yao Zhang}
is currently a Post-Doctoral Researcher with Northwestern Polytechnical University, Xi'an,
China. He received the Ph.D. degree in Telecommunication Engineering from Xidian University, Xi'an, China, in 2020. He was a Research Assistant and Post-Doctoral Fellow with The Hong Kong Polytechnic University in 2019 and 2021, respectively. His current research interests include edge intelligence, crowdsensing, and networked autonomous driving.
\end{IEEEbiographynophoto}
\begin{IEEEbiographynophoto}{Zhiwen Yu}
(Senior Member, IEEE) received the Ph.D. degree of engineering in computer science and technology from Northwestern Polytechnical
University, in 2005. He is currently a professor at Northwestern Polytechnical University, China. He has worked as a research fellow at the Academic Center for Computing and Media Studies, Kyoto University, Japan, from February 2007 to January 2009, and a postdoctoral researcher at the Information Technology Center, Nagoya University, Japan, in 2006-2007. He has been an Alexander von Humboldt fellow at Mannheim University, Germany, from November 2009 to October 2010. His research interests include pervasive computing, context-aware systems, human-computer interaction, mobile social networks, and personalization.
\end{IEEEbiographynophoto}
\begin{IEEEbiographynophoto}{Jun Zhang}
(Fellow, IEEE) received the B.Eng. degree in Electronic Engineering from the University of Science and Technology of China in 2004, the M.Phil. degree in Information Engineering from the Chinese University of Hong Kong in 2006, and the Ph.D. degree in Electrical
and Computer Engineering from the University of Texas at Austin in 2009. He is an Associate
Professor in the Department of Electronic and Computer Engineering at the Hong Kong University of Science and Technology. His research interests include wireless communications and networking, mobile edge computing and edge AI, and cooperative AI.
Dr. Zhang co-authored the book Fundamentals of LTE (Prentice-Hall, 2010). He is a co-recipient of several best paper awards, including the 2021 Best Survey Paper Award of the IEEE Communications Society, the 2019 IEEE Communications Society \& Information Theory Society Joint Paper Award, and the 2016 Marconi Prize Paper Award in Wireless Communications. Two papers he co-authored received the Young Author Best Paper Award of the IEEE Signal Processing Society in 2016 and 2018, respectively. He also received the 2016 IEEE ComSoc Asia-Pacific Best Young Researcher Award. He is an Editor of IEEE Transactions on Communications, IEEE Transactions on Machine Learning in Communications and Networking, and was an editor of IEEE Transactions on Wireless Communications (2015-2020). He served as a MAC track co-chair for IEEE Wireless Communications and Networking Conference (WCNC) 2011 and a co-chair for the Wireless Communications Symposium of IEEE International Conference on Communications (ICC) 2021. He is an IEEE Fellow and an IEEE ComSoc Distinguished Lecturer.
\end{IEEEbiographynophoto}
\begin{IEEEbiographynophoto}{Liang Wang}
received the Ph.D. degree in computer science from the Shenyang Institute of Automation (SIA), Chinese Academy of Sciences, Shenyang, China, in 2014. He is currently an associate professor at Northwestern Polytechnical University, Xi'an, China. His research interests include ubiquitous computing, mobile crowd sensing, and data mining.
\end{IEEEbiographynophoto}
\begin{IEEEbiographynophoto}{Tom H. Luan} 
(Senior Member, IEEE) received the B.E. degree from the Xi'an Jiaotong University, China,
the Master degree from the Hong Kong University of Science and Technology, Hong Kong, and the Ph.D. degree from the University of Waterloo, Canada, all in Electrical and Computer Engineering. During 2013 to 2017, Dr. Luan was a Lecturer in Mobile and Apps at the Deakin University, Australia. Since 2017, he is with the School of Cyber Engineering in Xidian University, China, as a professor. His research mainly focuses on the content distribution and media streaming in vehicular ad hoc networks and peer-to-peer networking, and protocol design and performance evaluation of digital network and edge computing. Dr. Luan has published over 150 peer reviewed papers in journal and conferences, including IEEE TON, TMC, TMM, TVT and Infocom. He won the 2017 IEEE VTS Best Land Transportation Best Paper award, and IEEE ICCS 2018 best paper award.
\end{IEEEbiographynophoto}
\begin{IEEEbiographynophoto}{Bin Guo}
(Senior Member, IEEE) received the PhD degree in computer science from Keio University, Minato, Japan, in 2009. He is currently a professor with Northwestern Polytechnical University, Xi'an, China. He was a postdoctoral researcher with the Institut TELECOM SudParis, France. His research interests include ubiquitous computing, mobile crowd sensing, and human-computer interaction.
\end{IEEEbiographynophoto}
\begin{IEEEbiographynophoto}{Chau Yuen}
(Fellow, IEEE) received his B.Eng. and Ph.D. from Nanyang Technological University (NTU), Singapore, in 2000 and 2004, respectively. He was a Post-Doctoral Fellow with Lucent Technologies Bell Labs, Murray Hill, in 2005, and a Visiting Assistant Professor with The Hong Kong Polytechnic University in 2008. From 2006 to 2010, he was with the Institute for Infocomm Research (12R), Singapore. From 2010 to March of 2023, he was with the Singapore University of Technology and Design. Since April 2023, he has been with the School of EEE, NTU.

\end{IEEEbiographynophoto}
\end{document}